\documentclass[10pt, a4paper]{article}

\usepackage[final]{lrec2026} %
\usepackage[utf8]{inputenc} %
\usepackage[T1]{fontenc}    %
\usepackage{hyperref}       %
\usepackage{url}            %
\usepackage{booktabs}       %
\usepackage{multirow}
\usepackage{amsfonts}       %
\usepackage{nicefrac}       %
\usepackage{microtype}      %
\usepackage{xcolor}         %
\usepackage{enumitem}
\usepackage{nicematrix}
\usepackage{adjustbox}
\usepackage{comment}
\usepackage{listings}
\usepackage{pifont}
\usepackage{longtable}
\usepackage{titlesec}
\usepackage[table]{xcolor}
\usepackage{float}
\usepackage[most]{tcolorbox}
\newtcolorbox[use counter=section]{promptbox}[2][]{%
  colback=gray!5!white, colframe=gray!75!black,
  fonttitle=\bfseries,
  title=Prompt~\thesection: #2,#1
}
\definecolor{myred}{RGB}{206, 50, 16}
\definecolor{mygreen}{RGB}{20, 157, 51}

\newcommand{\deltaneg}[1]{\scriptsize{\bf \textcolor{myred}{(-{#1})}}}
\newcommand{\deltapos}[1]{\scriptsize{\bf \textcolor{mygreen}{(+{#1})}}}

\definecolor{mylightblue}{HTML}{e0ecee}
\definecolor{myblue}{HTML}{2ea4bd}

\title{Multimodal LLMs Do Not Compose Skills Optimally Across Modalities}

\name{Paula Ontalvilla, Aitor Ormazabal, Gorka Azkune} 

\address{HiTZ Center - Ixa, University of the Basque Country (EHU) \\
         \{paula.ontalvilla, aitor.ormazabal, gorka.azcune\}@ehu.eus\\}

\abstract{
  Skill composition is the ability to combine previously learned skills to solve new tasks. As neural networks acquire increasingly complex skills during their pretraining, it is not clear how successfully they can compose them. In this paper, we focus on Multimodal Large Language Models (MLLM), and study their ability to compose skills across modalities. To this end, we design three evaluation tasks which can be solved sequentially composing two modality-dependent skills, and evaluate several open MLLMs under two main settings: i) prompting the model to directly solve the task, and ii) using a two-step cascaded inference approach, which manually enforces the composition of the two skills for a given task. Even with these straightforward compositions, we find that all evaluated MLLMs exhibit a significant \textbf{cross-modality skill composition gap}. To mitigate the aforementioned gap, we explore two alternatives: i) use chain-of-thought prompting to explicitly instruct MLLMs for skill composition and ii) a specific fine-tuning recipe to promote skill composition. %
  Although those strategies improve model performance, they still exhibit significant skill composition gaps, suggesting that more research is needed to improve cross-modal skill composition in MLLMs.  \\ \newline  }

\begin{document}

\maketitleabstract

\section{Introduction}

Multimodal Large Language Models (MLLM) have become increasingly capable in recent years, receiving ever-growing interest from the research community and industry \citep{alayrac2022flamingovisuallanguagemodel, driess2023palmeembodiedmultimodallanguage, liu2023visualinstructiontuning, openai2024gpt4technicalreport}. However, even as these models become capable of solving more and more complex tasks, the mechanisms through which they do so remain unclear, spawning extensive research into their internal workings \citep{Dang2024ExplainableAI}.

MLLMs are often trained using a capable LLM as a starting point, and visual or other multimodal capabilities are added by further fine-tuning it on modality-specific tasks. While some work has studied the effects of this type of training on the original model's textual capabilities \citep{Lin2023VILAOP,dai2024nvlmopenfrontierclassmultimodal}, the resulting MLLM's ability to coherently combine its visual and textual skills is not well understood. In this work, we pose the following question: \textit{Can MLLMs successfully compose skills across modalities?} 

More specifically, we devise 3 image-to-text tasks that can be solved by composing a visual skill with a textual one, in a way that would be trivial for a human: 1) Reasoning over questions rendered in an image (OCR + textual reasoning), 2) Object counting (object identification + counting), and 3) Card playing (card identification + textual reasoning over the game state). By comparing the model's natural performance on these tasks with a cascaded inference approach that forces skill composition by making two calls to the same model, we can measure a ``cross-modality skill composition gap''. We measure this gap on a wide array of popular open-source MLLMs and datasets and find that, strikingly, \textbf{all tested models exhibit a significant cross-modality skill composition gap}, even when the required combination is straightforward\footnote{\url{https://github.com/paulaonta/skillComposition}}.

Having demonstrated the existence of this gap, we explore two possible mitigation strategies: one leveraging a prompt specifically constructed to induce the relevant skill composition and a second approach exploring fine-tuning to induce cross-modal skill composition. We show that the specially constructed prompt partially reduces the skill composition gap, but it suffers from lack of scalability as it requires a task-specific prompt. Fine-tuning also improves the results and it contributes to generalization to some extent, yet the gap remains. %

Our results show that MLLMs do not optimally combine their visual and textual skills even in very straightforward scenarios, prompting further research into the root cause and effective mitigation of this phenomenon.

\begin{figure*}
  \centering
  \includegraphics[scale=0.22]{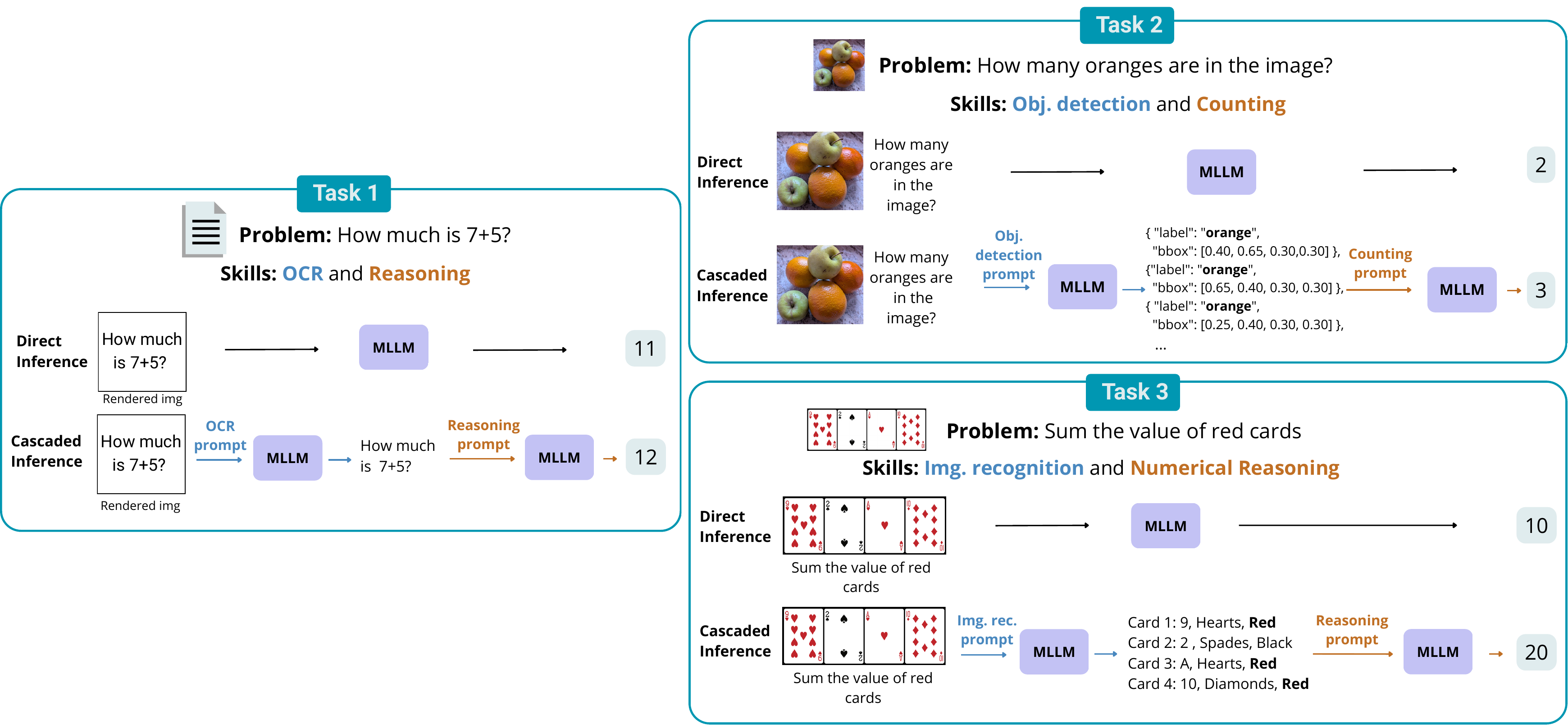}  
  \caption{%
  We define three tasks that can be solved with a trivial sequential composition of two modality-specific skills. We evaluate MLLMs on those tasks using direct inference, i.e. standard inference. We compare the performance with the cascaded inference, where we manually induce the required skill composition with two calls to the MLLM.}
  \label{fig:mainfigure}
\end{figure*}

\section{Related Work}\subsection{Skill composition in neural networks}
There are different ways to understand and approach \textbf{skills} and \textbf{skill composition} in the literature, both in robotics \cite{qureshicomposing, liu2025skill} and in LLMs \cite{zhao2024can}. We will cover the literature of the domains that are closest to our work.

In the artificial agent community, skills are defined as separated atomic tools that an agent can use at inference time via API calls \cite{zheng2025skillweaverwebagentsselfimprove, NEURIPS2023_871ed095}. Thus, skill composition is the ability of those agents to compose the tools available to solve different tasks. To achieve skill composition \citet{zheng2025skillweaverwebagentsselfimprove} focus on self-improvement techniques proposing a structured pipeline that allows web agents to autonomously discover, refine, and recombine reusable skills over time. In contrast, \citet{NEURIPS2023_871ed095} rely on an LLM to plan a sequence of tools to be used to approach a given task.

The LLM community offers some other views on skill composition. \citet{zhao2024can} defines compositional generalization as ``the capacity to combine learned skills in novel ways that are not encountered during training''. We adopt this definition in our formulation of skill composition. They used the SkillMix dataset \cite{yu2023skill} to show that LLMs can learn to compose skills to some extent through specific fine-tuning. In this framework, the skills they consider are  \textit{literary, rhetorical, reasoning, logic, theory of mind, pragmatics, common sense} and \textit{physical knowledge}.

We also understand skills as specific abilities learned by neural networks during training. However, rather than focusing on diffuse skill categories and complex compositions for text-only LLMs as in \citet{yu2023skill}, we focus on the multimodal scenario, where visual and textual skills can be sharply divided, and study straightforward scenarios that can be solved by combining one \textbf{visual} skill with one \textbf{textual} skill. This allows us to measure the cross-modality skill composition gap, and study how well MLLMs are able to compose skills across modalities.

\subsection{Multimodal Large Language Models (MLLM)}

In this paper, we will focus on Multimodal Large Language Models (MLLMs) that receive visual and textual input and generate textual output. Following the current dominant paradigm, we consider open weights models that follow the late fusion paradigm, where a pretrained visual encoder and LLM are connected and further trained for vision-language instruction following. Examples of this paradigm are the LLaVA family \cite{liu2024improved}, Llama 3.2 \cite{grattafiori2024llama}, Molmo \cite{deitke2024molmo} and Qwen2.5-VL \cite{bai2025qwen2}.

Studying skill composition requires visibility into when and how individual skills are acquired during training. For this reason, we restrict our analysis to open-source MLLMs, whose training stages and objectives are publicly documented. Closed-source models are excluded, as their training data and process are not transparent, making it impossible to determine which skills they were explicitly trained for.

Open-source MLLMs offer an ideal test bench to study skill composition for two reasons. Firstly, modality-specific skills can be easily distinguished and tracked on their training stages. For example, current MLLMs are often trained to perform Optical Character Recognition (OCR), which we categorize as a visual skill (see \S\ref{sec:skills}). In contrast, modern pretrained text-only LLMs are commonly trained on data that supports arithmetic reasoning, a textual skill. Secondly, the dominant late-fusion paradigm adds a modality specific (e.g. visual) encoder later in training, raising questions of how well the final model is able to combine skills from both modalities.

\section{Skill Composition and Multimodal Large Language Models}
\label{sec:skills}

In this section, we describe how we understand skills and skill composition, and how we design our tasks to study cross-modality skill composition ability in  MLLMs.

A \textbf{skill} is the ability acquired by a model as a result of the training process. For instance, a model trained to describe images will acquire the skill of image captioning. In this paper, we distinguish between visual skills, that demand image processing and representation, and textual skills, that only operate on text. 

A \textbf{task} is a set of instructions and correct answers that are used to measure specific skills. Tasks and skills are closely related, since tasks are used to guide the MLLM training process. In the example before, if we use the task of image captioning to train an MLLM, the model will acquire the image captioning skill. However, the relation between skills and tasks is not always $1:1$, but $N:1$, meaning that, generally, a set of skills may be needed to solve a task. In that sense, we can define tasks that are not used in the training process but can be solved using the skills acquired in that training process. 

Based on these definitions, we will adhere to the definition of compositional generation of \cite{zhao2024can} and state that: \textbf{skill composition} is the capacity to combine learned skills in novel ways that are not encountered during training. 

In the context of MLLMs, the pre-trained backbone LLM will have acquired a set of textual skills, while the multimodal fine-tuning will imbue the model with visual skills. We aim to study whether trained MLLMs can successfully combine and compose these skills across modalities. To that end, we design three tasks that can be solved by straightforwardly composing one visual skill and one textual skill, as can be seen in Figure~\ref{fig:mainfigure}.

\paragraph{Task 1: Reasoning over rendered text. } Given a textual problem, which is rendered as an image, the model is asked (in textual form) to solve the problem which requires some sort of reasoning: mathematical, general knowledge or commonsense reasoning. Task 1 can be solved composing Optical Character Recognition (OCR), a visual skill, and language-based reasoning, a textual skill. 

\paragraph{Task 2: Object counting. } Given an image, the model is asked by text to count the occurrences of a specific object. This task can be solved composing the visual skill of object detection and the textual skill of counting a given string in a text. 

\paragraph{Task 3: Card playing. } Given an image containing four cards from a poker deck, we instruct the model to order cards or to perform summations under different conditions. To solve those tasks, the visual skill of image recognition and the textual skill of numerical reasoning can be composed.

\section{Experimental Settings} \label{sec:experimental_settings}
In this section, we describe the models we chose for our experimentation (\S\ref{ssec:models}), as  well as our approach for measuring the skill composition gap (\S\ref{ssec:method}).
\subsection{Models} \label{ssec:models}
We evaluate a range of popular open MLLMs with known training skills and tasks, specifically LLaVA 1.6-Vicuna-13B, LLaVA 1.6-Mistral-7B \cite{liu2024llavanext}, Llama 3.2 11B \cite{grattafiori2024llama}, Molmo-7B and 72B \cite{deitke2024molmo} and Qwen2.5-VL-7B and 72B \cite{bai2025qwen2}. We do not evaluate closed models, as their training recipes and their backbone LLMs are unknown, and consequently, we cannot identify the specific visual and textual skills they have been explicitly trained for. Based on the papers and technical reports corresponding to each chosen model, we verify that all the MLLMs we selected have the visual and textual skills required by the tasks described in \S\ref{sec:skills}, except for LLaVA and Llama models, which have not been explicitly trained for object detection.

\subsection{Measuring the skill composition gap} \label{ssec:method}

Our goal is to evaluate how well MLLMs can compose their visual and textual skills. To that end, we evaluate models on two different inference setups: (1) \texttt{Direct inference}, where the model is prompted with an image and an associated question to directly solve the task; (2) \texttt{Cascaded inference} which enforces skill composition by making two calls to the same model. Direct inference will show the natural behavior of the MLLMs, letting them solve the task without further instructions. In cascaded inference, on the other hand, we first ask the MLLM to use the relevant visual skill for the target task to produce an intermediate result. This result is then used as the input of the second step, where the MLLM is explicitly asked to use the relevant textual skill (see Figure \ref{fig:mainfigure}).

Based on these two setups, we define the \textbf{cross-modality skill composition gap}\footnote{For simplicity, we  use \textit{cross-modality skill composition gap} and \textit{skill composition gap} interchangeably throughout the paper.} of an MLLM as the difference of performance between cascaded and direct inferences. If the model's performance under direct inference is lower than the cascaded version, we can conclude that the model is not composing its own skills in an optimal way, since cascaded inference only uses the model itself to solve the individual skills.

Finally, as an upper bound for cascaded inference, we also provide the oracle results when possible. The \texttt{oracle inference} prompts the models to use the relevant textual skill for the target task, but using as input the gold standard result of the visual skill\footnote{For instance, for Task 1, \textit{reasoning over rendered text}, oracle would prompt the model with the full original textual problem statement, instead of using the model's own OCR result.}. Thus, the oracle inference setup provides the best results an MLLM could achieve with our designed skill composition.

When evaluating model generations, we distinguish between two cases: i) multiple-choice answers, where models output answers in letter forms and we compute exact-match accuracy; ii) open-ended generation, depending on the dataset, evaluated via regex or LLM-based methods as standard (see Appendix~\ref{app:implementation}). 
For both direct and cascaded inference approaches, we prompt the model to reason step-by-step. See Appendix~\ref{app:prompts} for the exact prompts used in each task and setup.  %

\section{Results} \label{sec:results} 

\subsection{Results for Task 1: Reasoning over rendered text}

As mentioned in \S\ref{sec:skills}, we design Task 1 to be solvable by composing the visual skill of Optical Character Recognition (OCR) with the textual skill of reasoning. Since we use open models, we know that all of them have been trained for both OCR and language-based reasoning. To build relevant datasets for Task 1, we render common text-only benchmarks which assess mathematical reasoning (GSM8K~\cite{cobbe2021gsm8k} and MATH~\cite{hendrycks2024measuring}, using the test
set from~\citet{lightman2023let}), commonsense reasoning (ARC Easy and Challenge~\cite{clark2018think}) and general knowledge and reasoning ability (MMLU~\cite{hendrycks2020measuring}). For all datasets, we render the plain text as images. An exception is MATH, which contains LaTeX-formatted content; so, we first compile it before rendering. Since we construct the visual dataset from an original textual one, we naturally have access to the gold results for the OCR task (visual skill) and we can provide results for oracle inference.

Table~\ref{tab:task1_1results} reports our results across all datasets under the three inference setups. As shown in the table \textbf{all tested MLLMs exhibit a skill composition gap}, as cascaded inference always outperforms direct inference. The gap ranges from  $22.19$ points in the case of LLaVA 1.6-Vicuna-13B, to $2.45$ points in the case of Qwen2.5-VL-7B. Interestingly, the gap for Qwen2.5-VL-72B is only $0.39$ points, which does not indicate a meaningful skill composition gap.

\begin{table*}[htb]
  \centering
  \begin{adjustbox}{ max height=5.5cm}
  \begin{tabular}{llcccccc}
    \toprule
     \textbf{Model} &  \textbf{Setup} & \multicolumn{5}{c}{\textbf{Datasets}} & \; \textbf{Avg} \\
     &  &   \textbf{GSM8K} & \textbf{ARC-E} & \textbf{ARC-C} & \textbf{MMLU} & \textbf{MATH} &   \\
    \midrule 
    \multirow{3}{*}{LLaVA 1.6-Vicuna-13B}& Oracle&    
    29.11&83.83&70.05&51.73&4.60& \;47.86\\ 
    \cmidrule(lr){3-8}
    & \cellcolor{mylightblue}Direct&\cellcolor{mylightblue}8.79&\cellcolor{mylightblue}28.82&\cellcolor{mylightblue}27.21&\cellcolor{mylightblue}26.64&\cellcolor{mylightblue}\textbf{3.20}& \cellcolor{mylightblue}\;18.93 \\
    & \cellcolor{mylightblue}Cascaded&  \cellcolor{mylightblue}\textbf{26.91}&\cellcolor{mylightblue}\textbf{72.64}&\cellcolor{mylightblue}\textbf{60.75}&\cellcolor{mylightblue}\textbf{43.68}&\cellcolor{mylightblue}1.60& \cellcolor{mylightblue}\;\textbf{41.12}\\
    \midrule
    \multirow{3}{*}{LLaVA 1.6-Mistral-7B}& Oracle&   42.53 & 87.42 & 75.77& 53.01& 12.80  & \;54.31 \\
      \cmidrule(lr){3-8}
    & \cellcolor{mylightblue}Direct& \cellcolor{mylightblue}11.75&\cellcolor{mylightblue}51.51&\cellcolor{mylightblue}45.05&\cellcolor{mylightblue}31.68&\cellcolor{mylightblue}5.60&\cellcolor{mylightblue}\;29.12\\
    & \cellcolor{mylightblue}Cascaded&   \cellcolor{mylightblue}\textbf{36.77} &\cellcolor{mylightblue}\textbf{75.71}&\cellcolor{mylightblue}\textbf{62.37}&\cellcolor{mylightblue}\textbf{46.49}&\cellcolor{mylightblue}\textbf{7.00}& \cellcolor{mylightblue}\;\textbf{45.67}\\
    \midrule
    \multirow{3}{*}{Llama 3.2 11B}& Oracle& 84.23&94.69&87.71&72.08&46.20  &\;76.98    \\
      \cmidrule(lr){3-8}
    & \cellcolor{mylightblue}Direct&  \cellcolor{mylightblue}63.59&\cellcolor{mylightblue}69.23&\cellcolor{mylightblue}61.43&\cellcolor{mylightblue}46.37&\cellcolor{mylightblue}36.40 &\cellcolor{mylightblue}\; 55.40 \\
    & \cellcolor{mylightblue}Cascaded& \cellcolor{mylightblue}\textbf{81.05}&\cellcolor{mylightblue}\textbf{94.74}&\cellcolor{mylightblue}\textbf{87.88}&\cellcolor{mylightblue}\textbf{68.21}&\cellcolor{mylightblue}\textbf{40.40} & \cellcolor{mylightblue}\; \textbf{74.46} \\  
    \midrule
    \multirow{3}{*}{Molmo-7B}& Oracle& 70.28 & 90.61 & 82.08 & 60.88 & 27.40&\;66.25  \\
      \cmidrule(lr){3-8}
    & \cellcolor{mylightblue}Direct& \cellcolor{mylightblue}60.27 & \cellcolor{mylightblue}77.10 & \cellcolor{mylightblue}66.89 & \cellcolor{mylightblue}45.07 & \cellcolor{mylightblue}16.00 &\cellcolor{mylightblue}\; 53.07 \\
    & \cellcolor{mylightblue}Cascaded&  \cellcolor{mylightblue}\textbf{66.87} & \cellcolor{mylightblue}\textbf{86.57} & \cellcolor{mylightblue}\textbf{76.02} & \cellcolor{mylightblue}\textbf{56.67} & \cellcolor{mylightblue}\textbf{23.00} & \cellcolor{mylightblue}\; \textbf{61.83} \\
    \midrule
    \multirow{3}{*}{Molmo-72B}& Oracle& 93.10 & 98.53 & 95.05& 81.93 & 62.40&\;86.20  \\
      \cmidrule(lr){3-8}
    & \cellcolor{mylightblue}Direct& \cellcolor{mylightblue}87.64 & \cellcolor{mylightblue}93.64 & \cellcolor{mylightblue}88.14 & \cellcolor{mylightblue}67.36 & \cellcolor{mylightblue}40.40 &\cellcolor{mylightblue}\; 75.44 \\
    & \cellcolor{mylightblue}Cascaded&  \cellcolor{mylightblue}\textbf{92.27} & \cellcolor{mylightblue}\textbf{97.81} & \cellcolor{mylightblue}\textbf{93.69} & \cellcolor{mylightblue}\textbf{75.61} & \cellcolor{mylightblue}\textbf{45.00} & \cellcolor{mylightblue}\; \textbf{80.88} \\
    \midrule
    \multirow{3}{*}{Qwen2.5-VL-7B}& Oracle&  85.14&96.21&88.82&71.32&65.60 & \;81.42 \\
      \cmidrule(lr){3-8}
    & \cellcolor{mylightblue}Direct& \cellcolor{mylightblue}79.45&\cellcolor{mylightblue}94.78&\cellcolor{mylightblue}86.17&\cellcolor{mylightblue}66.06&\cellcolor{mylightblue}\textbf{61.80} &\cellcolor{mylightblue}\;77.65 \\
    & \cellcolor{mylightblue}Cascaded&  \cellcolor{mylightblue}\textbf{83.85}& \cellcolor{mylightblue}\textbf{96.00}& \cellcolor{mylightblue}\textbf{88.57}& \cellcolor{mylightblue}\textbf{70.28}&\cellcolor{mylightblue}\textbf{61.80} &\cellcolor{mylightblue}\;\textbf{80.10} \\
     \midrule
    \multirow{3}{*}{Qwen2.5-VL-72B}& Oracle&  95.75&99.03&96.67&85.55&80.40 & \;91.48\\
      \cmidrule(lr){3-8}
    & \cellcolor{mylightblue}Direct& \cellcolor{mylightblue}95.22&\cellcolor{mylightblue}98.48&\cellcolor{mylightblue}95.39&\cellcolor{mylightblue}80.94&\cellcolor{mylightblue}\textbf{77.20} &\cellcolor{mylightblue}\;89.45 \\
    & \cellcolor{mylightblue}Cascaded&  \cellcolor{mylightblue}\textbf{95.30}& \cellcolor{mylightblue}\textbf{98.57}& \cellcolor{mylightblue}\textbf{96.59}& \cellcolor{mylightblue}\textbf{82.94}&\cellcolor{mylightblue}75.80 &\cellcolor{mylightblue}\;\textbf{89.84} \\
    \bottomrule
  \end{tabular}
  \end{adjustbox}
    \caption{Exact-match accuracy of MLLMs for Task 1 (reasoning over rendered text) under three different inference setups on each benchmark. Oracle is provided as an upper bound reference for the cascaded inference. The best performance for each model under the two main inference setups (cascaded and direct) is shown in \textbf{bold}. The difference between the cascaded and direct results reveals the skill composition gap, which is significant for almost all MLLMs.}
    \label{tab:task1_1results}
\end{table*}

While every model exhibits a cross-modality skill composition gap, both small and large models of the Molmo and Qwen2.5-VL families exhibit significantly smaller gaps than other models. Manual inspection of the \textit{Pixmo Ask Model Anything  }\cite{deitke2024molmo} dataset used to train Molmo shows examples of math problems presented in images, which  provides explicit training data for the skill composition we are testing in this task. For Qwen2.5-VL \cite{bai2025qwen2}, the technical report also mentions the usage of invoices, forms, receipts, table/diagram corpora for training, making the model naturally reason over rendered textual information. This stresses the importance of knowing the training recipes of the models we test, as they allow us to track capabilities to training data, and distinguish ``novel'' skill composition from skills a model has explicitly been trained on.

Furthermore,  the results for the oracle setup are significantly higher than the cascaded ones, even for the most capable models, meaning that the MLLM's OCR errors account for substantial degradation in final performance, which is further detailed in Appendix~\ref{app:task1}.

\subsection{Results for Task 2: Object counting}

Object counting can be decomposed into two steps: object detection or pointing (a visual skill) and counting the occurrences of a specific string in a text (a textual skill). To implement Task 2, we use the CV-Bench~\cite{tong2024cambrian} benchmark, which, given an image, prompts the model to count a specific type of object. As CV-Bench does not provide the object annotations of the images, we can neither evaluate MLLMs with oracle inference, nor individually analyze the visual and textual skills of the models. To better study these aspects, we introduce a new dataset: COCO-Count. This dataset is built from the 2017 validation split of COCO~\cite{lin2014microsoft}, and consists of images paired with annotated bounding boxes and points and the corresponding number of instances of a target object (see Appendix~\ref{app:coco} for details). %

\begin{table}[htb]
  \centering
   \begin{adjustbox}{max width=\columnwidth}
  \begin{tabular}{llcccc}
    \toprule    
     \textbf{Model} &  \textbf{Setup} & \multicolumn{2}{c}{\textbf{Datasets}} & \; \textbf{Avg}   \\
     & &   \textbf{CV-Bench} & \textbf{COCO-Count} & \\
    \midrule 
    \multirow{3}{*}{Molmo-7B}&  Oracle & N/A & 38.11 & \;N/A \\
    \cmidrule(lr){3-5}
    & \cellcolor{mylightblue}Direct& \cellcolor{mylightblue}\textbf{60.53} & \cellcolor{mylightblue} \textbf{56.62} & \cellcolor{mylightblue}\;\textbf{58.58}\\
    & \cellcolor{mylightblue}Cascaded& \cellcolor{mylightblue}33.12  & \cellcolor{mylightblue}16.07 &\cellcolor{mylightblue}\;24.60 \\
    \midrule
    \multirow{3}{*}{Molmo-72B}&  Oracle & N/A &79.68& \;N/A \\
    \cmidrule(lr){3-5}
    & \cellcolor{mylightblue}Direct& \cellcolor{mylightblue}64.47 & \cellcolor{mylightblue}\textbf{57.53}& \cellcolor{mylightblue}\;\textbf{61.00}\\
    & \cellcolor{mylightblue}Cascaded& \cellcolor{mylightblue}\textbf{65.61 } & \cellcolor{mylightblue}53.29 &\cellcolor{mylightblue}\;59.45 \\
    \midrule
    \multirow{3}{*}{Qwen2.5-VL-7B} & Oracle & N/A &  99.53 &\; N/A\\
    \cmidrule(lr){3-5}
    & \cellcolor{mylightblue}Direct& \cellcolor{mylightblue}65.48 & \cellcolor{mylightblue}66.36&\cellcolor{mylightblue}\;65.92  \\
    & \cellcolor{mylightblue}Cascaded&  \cellcolor{mylightblue}\textbf{70.68} & \cellcolor{mylightblue}\textbf{70.31} &\cellcolor{mylightblue}\;\textbf{70.50} \\
    \midrule
    \multirow{3}{*}{Qwen2.5-VL-72B} & Oracle & N/A &  99.91 &\; N/A\\
    \cmidrule(lr){3-5}
    & \cellcolor{mylightblue}Direct& \cellcolor{mylightblue}69.16 & \cellcolor{mylightblue}65.17&\cellcolor{mylightblue}\;67.17  \\
    & \cellcolor{mylightblue}Cascaded&  \cellcolor{mylightblue}\textbf{70.81} & \cellcolor{mylightblue}\textbf{70.44} &\cellcolor{mylightblue}\;\textbf{70.63} \\
    \bottomrule
  \end{tabular}
   \end{adjustbox}
    \caption{Exact-match accuracy of MLLMs for Task 2 (object counting) under three different inference setups on each benchmark. Oracle is provided as an upper bound reference for the cascaded inference (when possible). The best performance for each model under the two main inference setups (cascaded and direct) is shown in \textbf{bold}. The difference between the cascaded and direct results reveals the skill composition gap, which is significant for both Qwen2.5-VL models, but non-existent for both Molmo models (see text for details).}
    \label{tab:task2_1results}
\end{table}

Results for each benchmark can be seen in Table~\ref{tab:task2_1results}. We only show the results of models that have been trained for the required visual and textual skills (as discussed in \S\ref{ssec:models} only Molmo and Qwen2.5-VL have the object detection skill). Results for the remained models can be found in Appendix~\ref{app:task2}. We find that Qwen2.5-VL models exhibit a clear skill composition gap, whereas Molmo models do not. To investigate this further, we measured the performance of each model on each individual skill to determine whether the absence of a composition gap is due to effective skill integration, or simply because the model fails to perform one or both skills correctly. We use the COCO-Count dataset for this analysis, as it is the only dataset that includes annotated bounding boxes and points, which are necessary for assessing the models' counting ability in isolation.

Table~\ref{tab:task2_2results} summarizes these results. It can be seen that although Molmo has been reported to be trained on both skills, Molmo-7B shows poor performance in each one individually with the accuracy below 50\%. Molmo-72B performs better; however, its accuracy remains below 80\%, which can still be considered low, especially for the string counting textual skill, where Qwen2.5-VL models achieve nearly perfect performance. Therefore, Molmo models' low score in the cascaded setup cannot be attributed to a lack of skill composition, but rather to insufficient competence in the individual skills themselves.

\begin{table}[htb]
  \centering
  \begin{adjustbox}{max width=\columnwidth}
  \begin{tabular}{llc}
    \toprule    
     \textbf{Model} &  \textbf{Skill} & \textbf{COCO-Count}   \\
    \midrule 
    \multirow{3}{*}{Molmo-7B}& Obj. detection (visual) & 46.82 \\
    & Str. counting (textual) & 38.11  \\    
     & Cascaded & 16.07 \\
    \midrule
     \multirow{3}{*}{Molmo-72B}& Obj. detection (visual) & 65.87 \\
    & Str. counting (textual) & 79.68  \\    
     & Cascaded & 53.29 \\
     \midrule
    \multirow{3}{*}{Qwen2.5-VL-7B} & Obj. detection (visual)& 67.65  \\
    & Str. counting (textual) & 99.53 \\    
    & Cascaded & 70.31  \\
         \midrule
    \multirow{3}{*}{Qwen2.5-VL-72B} & Obj. detection (visual)& 70.76  \\
    & Str. counting (textual) & 99.91 \\    
    & Cascaded & 70.44  \\
    \bottomrule
  \end{tabular}
  \end{adjustbox}
      \caption{Accuracy of the MLLMs for the visual skill of object detection and the textual skill of string counting, which are involved in Task 2 (object counting). We also provide the cascaded result as reference, as the combination of both skills.}
  \label{tab:task2_2results}
\end{table}      

In summary, we again observe a clear skill composition gap for both Qwen2.5-VL models. This gap is not evident for the Molmo models, which perform poorly on the individual skills themselves. Further analysis reveals that both Molmo models struggle with the simple skill of textual counting, likely contributing to their weak performance in the cascaded setting. Interestingly, Molmo models perform comparably well under direct inference, potentially because they were directly trained on this task.

\subsection{Results for Task 3: Card playing}

Task 3 is formulated as the sequential composition of image recognition (visual skill) and numerical reasoning (textual skill). To this end, we design two different card games similar to \citet{chu2025sftmemorizesrlgeneralizes}. The first card game presents a sequence of four randomly ordered poker cards. The task is to determine the correct position of a new card such that the ascending order of the sequence is preserved. The second game also consists of four randomly selected cards, but unordered. The objective is to compute the sum of the card values according to specific conditions, where not only the card numbers but also the suits and colors play a crucial role in solving the task (details in Appendix~\ref{app:cards}). %

Table~\ref{tab:task3_1results} presents the results for this task, where the \textit{Sort} column reports performance on the first game and the \textit{Sum} column corresponds to the second game. As in previous tasks, we observe a skill composition gap for every model and game (with the exception LLaVA 1.6-Mistral-7B on Sum, where performance is very poor for both inference setups). Notably, this gap is quite pronounced even for models that showed a smaller gap in Task 1, such as Molmo-7B, Molmo-72B and Qwen2.5-VL-7B. We hypothesize that this may be due to the novelty of this task, which makes it less likely that these models were directly trained on it. However, in the case of Qwen2.5-VL-72B the gap is not very big suggesting that this model is better at composing the required skills for this task.

\begin{table}[t]
  \centering
  \begin{adjustbox}{max width=\columnwidth}
  \begin{tabular}{llccc}
    \toprule    
     \textbf{Model} & \textbf{Setup} & \multicolumn{2}{c}{\textbf{Datasets}} & \;\textbf{Avg} \\
      & & \textbf{Sort} & \textbf{Sum} &  \\
    \midrule 
    \multirow{3}{*}{LLaVA 1.6-Vicuna-13B} & Oracle    & 22.90 & 27.70 & \;25.30 \\\cmidrule(lr){3-5}
     & \cellcolor{mylightblue}Direct    & \cellcolor{mylightblue}21.10 & \cellcolor{mylightblue}2.80 & \cellcolor{mylightblue}\;11.95 \\
     & \cellcolor{mylightblue}Cascaded  & \cellcolor{mylightblue}\textbf{21.60} & \cellcolor{mylightblue}\textbf{3.80}  & \cellcolor{mylightblue}\;\textbf{12.70} \\
    \midrule
    \multirow{3}{*}{LLaVA 1.6-Mistral-7B}& Oracle    & 45.20 & 38.90 & \;42.05 \\ \cmidrule(lr){3-5}
     & \cellcolor{mylightblue}Direct    & \cellcolor{mylightblue}30.70 & \cellcolor{mylightblue}\textbf{4.40}  & \cellcolor{mylightblue}\;17.55 \\
     & \cellcolor{mylightblue}Cascaded  & \cellcolor{mylightblue}\textbf{39.60} & \cellcolor{mylightblue}3.90 & \cellcolor{mylightblue}\;\textbf{21.75} \\
    \midrule
    \multirow{3}{*}{Llama 3.2 11B}       & Oracle    & 59.60 & 62.10 & \;60.85 \\ \cmidrule(lr){3-5}
     & \cellcolor{mylightblue}Direct    & \cellcolor{mylightblue}44.60 & \cellcolor{mylightblue}11.30 & \cellcolor{mylightblue}\;27.95 \\
     & \cellcolor{mylightblue}Cascaded  & \cellcolor{mylightblue}\textbf{57.60} & \cellcolor{mylightblue}\textbf{44.20} & \cellcolor{mylightblue}\;\textbf{50.90} \\
    \midrule
    \multirow{3}{*}{Molmo-7B}           & Oracle    & 53.60 & 52.30 & \;52.95 \\ \cmidrule(lr){3-5}
     & \cellcolor{mylightblue}Direct    & \cellcolor{mylightblue}47.30 & \cellcolor{mylightblue}30.30 & \cellcolor{mylightblue}\;38.80 \\
     & \cellcolor{mylightblue}Cascaded  & \cellcolor{mylightblue}\textbf{49.80} & \cellcolor{mylightblue}\textbf{43.80} & \cellcolor{mylightblue}\;\textbf{46.80} \\
    \midrule
    \multirow{3}{*}{Molmo-72B}           & Oracle    & 93.30& 75.90 & \;84.60\\ \cmidrule(lr){3-5}
     & \cellcolor{mylightblue}Direct    & \cellcolor{mylightblue}87.90 & \cellcolor{mylightblue}70.80 & \cellcolor{mylightblue}\;79.35 \\
     & \cellcolor{mylightblue}Cascaded  & \cellcolor{mylightblue}\textbf{92.70} & \cellcolor{mylightblue}\textbf{77.90} & \cellcolor{mylightblue}\;\textbf{85.30} \\
    \midrule
    \multirow{3}{*}{Qwen2.5-VL-7B}      & Oracle    & 75.00 & 52.60 & \;63.80 \\ \cmidrule(lr){3-5}
     & \cellcolor{mylightblue}Direct    & \cellcolor{mylightblue}57.60 & \cellcolor{mylightblue}51.80 & \cellcolor{mylightblue}\;54.70 \\
     & \cellcolor{mylightblue}Cascaded  & \cellcolor{mylightblue}\textbf{74.40} & \cellcolor{mylightblue}\textbf{52.80} & \cellcolor{mylightblue}\;\textbf{63.60} \\
     \midrule\;
    \multirow{3}{*}{Qwen2.5-VL-72B}      & Oracle    & 97.30 & 99.70 & \;98.50 \\ \cmidrule(lr){3-5}
     & \cellcolor{mylightblue}Direct    & \cellcolor{mylightblue}96.00 & \cellcolor{mylightblue}99.40 & \cellcolor{mylightblue}\;97.70 \\
     & \cellcolor{mylightblue}Cascaded  & \cellcolor{mylightblue}\textbf{97.20} & \cellcolor{mylightblue}\textbf{99.60} & \cellcolor{mylightblue}\;\textbf{98.40} \\
    \bottomrule
  \end{tabular}
   \end{adjustbox}
   \caption{Exact-match accuracy of MLLMs for Task 3  under three different inference setups on each benchmark. Oracle serves as an upper bound for cascaded inference. The best results for each model under the direct and cascaded inference setups are shown in \textbf{bold}. The difference between the cascaded and direct results reveals the skill composition gap, which is significant for all MLLMs.}
  \label{tab:task3_1results}
\end{table}

Furthermore, a gap also exists between the oracle and direct setups. For some models, such as Qwen2.5-VL-7B and Qwen2.5-VL-72B, this gap is minimal, indicating that the limiting factor is its textual reasoning capability. %
We further experiment on the individual skill performance in Appendix~\ref{app:task3}, confirming visual understanding for this task is perfect in the case of both Qwen2.5-VL models.

\section{Mitigation strategies} \label{sec:mitigation}
Our results so far reveal a consistent lack of optimal skill composition across tasks. This observation naturally leads us to the question of \textit{ How can we mitigate this gap?} To address this, we explore two strategies. In \S\ref{ssec:CoT}, we introduce an approach based on Composition-specific Chain-of-Thought (CoT) prompting. In \S\ref{ssec:finetune}, we explore a fine-tuning strategy to enhance skill composition. %

\subsection{Composition-specific Chain-of-Thought prompting}
\label{ssec:CoT}

As the first mitigation strategy, we opted for Chain-of-Thought (CoT) prompting \cite{wei2022chain}. Specifically, we provide the model with an image and a CoT prompt that explicitly states which skills to combine and in what order (see Appendix~\ref{app:prompts} for the evaluated prompts). The underlying idea is that this approach should work in much the same way as the cascaded setup, since both aim to sequentially apply different skills, except that in this case the entire reasoning process is guided within a single prompt and a single model call. 

Table \ref{tab:CoT} reports the results for CoT prompting alongside the previous inference setups. The effectiveness of this strategy appears to be model-dependent. For example, Qwen2.5-VL-7B, Qwen2.5-VL-72B, Molmo-7B and Molmo-72B show consistent improvements across most datasets, outperforming the direct setup. Similarly, both LLaVA models mostly outperform the direct setup, but their gain is overall more significant. For instance, there is an improvement of $36.33$ points for LLaVA-Vicuna and $23.11$ points for LLaVA-Mistral on the ARC-Easy dataset. In contrast, Llama fails to make any meaningful gains, and tends to perform at or below the direct setup.

\begin{table*}[htb]
    \centering
  \small
  \begin{adjustbox}{ max width=\textwidth}
  \begin{tabular}{llccccc cc cc c}
    \toprule     
      \textbf{Model} & \textbf{Setup}& \multicolumn{5}{c}{\textbf{Task 1}} & \multicolumn{2}{c}{\textbf{Task 2}} & \multicolumn{2}{c}{\textbf{Task 3}} & \textbf{Avg}\\
      \cmidrule(rl){3-7} 
      \cmidrule(rl){8-9} 
      \cmidrule(rl){10-11}
      &   &  \textbf{GSM8K} & \textbf{ARC-E} & \textbf{ARC-C} & \textbf{MMLU} & \textbf{MATH} &\textbf{CV-B} &\textbf{COCO} &  \textbf{Sort}  & \textbf{Sum} & \\
    \midrule 
     \multirow{3}{*}{LLaVA 1.6-Vicuna-13B}
    & Direct & 8.79 & 28.82 & 27.21 & 26.64 & \underline{3.20}&  \multirow{3}{*}{N/A}& \multirow{3}{*}{N/A}&\underline{21.10}&2.80 & \;16.94 \\
    & Cascaded & \textbf{26.91} & \textbf{72.64} & \textbf{60.75} & \textbf{43.68} & 1.60&&&\textbf{21.60}&\underline{3.80} & \;\textbf{33.00} \\
    & \cellcolor{mylightblue}CoT 
      & \cellcolor{mylightblue}\underline{12.28} 
      & \cellcolor{mylightblue}\underline{65.15} 
      & \cellcolor{mylightblue}\underline{52.04} 
      & \cellcolor{mylightblue}\underline{37.76} 
      & \cellcolor{mylightblue}\textbf{6.60}
      & 
      &
      & \cellcolor{mylightblue}18.40      & \cellcolor{mylightblue}\textbf{3.90} & \;\cellcolor{mylightblue}\underline{28.02} \\
    \midrule
    \multirow{3}{*}{LLaVA 1.6-Mistral-7B}
    & Direct & 11.75 & 51.51 & 45.05 & 31.68 & 5.60&\multirow{3}{*}{N/A}&\multirow{3}{*}{N/A}&30.70&\textbf{4.40} & \;25.81 \\
    & Cascaded & \textbf{36.77} & \textbf{75.71} & \textbf{62.37} & \textbf{46.49} & \underline{7.00} &&&\textbf{39.60}&\underline{3.90} & \;\textbf{38.83}\\
    & \cellcolor{mylightblue}CoT 
      & \cellcolor{mylightblue}\underline{29.41} 
      & \cellcolor{mylightblue}\underline{74.62} 
      & \cellcolor{mylightblue}\underline{60.84} 
      & \cellcolor{mylightblue}\underline{42.51} 
      & \cellcolor{mylightblue}\textbf{7.60}
      & 
      &
      & \cellcolor{mylightblue}\underline{39.40}
      & \cellcolor{mylightblue}\textbf{4.40} & \;\cellcolor{mylightblue}\underline{36.97} \\
    \midrule
    \multirow{3}{*}{Llama 3.2 11B}
    & Direct & 63.59 & \underline{69.23} & \underline{61.43} & 46.37 & \underline{36.40}&\multirow{3}{*}{N/A}&\multirow{3}{*}{N/A}&44.60&11.30 & \;47.56 \\
    & Cascaded & \textbf{81.05} & \textbf{94.74} & \textbf{87.88} & \textbf{68.21} & \textbf{40.40}&&&\textbf{57.60}&\textbf{44.20} & \;\textbf{67.73} \\
    & \cellcolor{mylightblue}CoT 
      & \cellcolor{mylightblue}\underline{68.54} 
      & \cellcolor{mylightblue}68.18 
      & \cellcolor{mylightblue}59.64 
      & \cellcolor{mylightblue}\underline{48.23} 
      & \cellcolor{mylightblue}35.20
      & 
      &
      & \cellcolor{mylightblue}\underline{47.30}
      & \cellcolor{mylightblue}\underline{13.10} & \;\cellcolor{mylightblue}\underline{48.60} \\
    \midrule
    \multirow{3}{*}{Molmo-7B}
    & Direct & 60.27 & 77.10 & 66.89 & 45.07 & \underline{16.00}&\textbf{60.53}&\textbf{56.62}&\underline{47.30}&30.30 & \;\underline{51.12} \\
    & Cascaded & \textbf{66.87} & \textbf{86.57} & \textbf{76.02} & \textbf{56.67} & \textbf{23.00}&33.12&16.07&\textbf{49.80}&\textbf{43.80} & \;\textbf{50.21} \\
    & \cellcolor{mylightblue}CoT 
      & \cellcolor{mylightblue}\underline{62.70} 
      & \cellcolor{mylightblue}\underline{79.37} 
      & \cellcolor{mylightblue}\underline{68.08} 
      & \cellcolor{mylightblue}\underline{46.14} 
      & \cellcolor{mylightblue}14.00 
      & \cellcolor{mylightblue}\underline{48.22}
      & \cellcolor{mylightblue}\underline{50.00}
      & \cellcolor{mylightblue}27.50
      & \cellcolor{mylightblue}\underline{40.10} & \;\cellcolor{mylightblue}48.46 \\
    \midrule
        \multirow{3}{*}{Molmo-72B}
    & Direct &  87.64 & 93.64 & 88.14 & 67.36 & \underline{40.40} & 64.47 &\underline{57.53} & \underline{87.90} &\underline{70.80}& \;73.10  \\
    & Cascaded & \textbf{92.27} & \textbf{97.81} & \textbf{93.69} & \textbf{75.61} & \textbf{45.00} &\underline{65.61} &53.29 &\textbf{92.70} & \textbf{77.90} & \;\textbf{77.10 }\\
    & \cellcolor{mylightblue}CoT 
      & \cellcolor{mylightblue}\underline{91.96}
      & \cellcolor{mylightblue}\underline{95.79}
      & \cellcolor{mylightblue}\underline{89.33}
      & \cellcolor{mylightblue}\underline{68.49}
      & \cellcolor{mylightblue}\textbf{45.00} 
      & \cellcolor{mylightblue}\textbf{68.40}
      & \cellcolor{mylightblue}\textbf{67.51}
      & \cellcolor{mylightblue}85.80
      & \cellcolor{mylightblue}67.20 & \cellcolor{mylightblue}\;\underline{75.50} \\
    \midrule
    \multirow{3}{*}{Qwen2.5-VL-7B}
    & Direct & \underline{79.45} & 94.78 & 86.17 & 66.06 & \textbf{61.80} &65.48&\underline{66.36}&57.60&51.80 & \;69.94 \\
    & Cascaded & \textbf{83.85} & \textbf{96.00} & \textbf{88.57} & \textbf{70.28} & \textbf{61.80}&\textbf{70.68}&\textbf{70.31}&\textbf{74.40}&\underline{52.80} & \;\textbf{74.30} \\
    & \cellcolor{mylightblue}CoT 
      & \cellcolor{mylightblue}79.38 
      & \cellcolor{mylightblue}\underline{95.37} 
      & \cellcolor{mylightblue}\underline{86.68} 
      & \cellcolor{mylightblue}\underline{67.37} 
      & \cellcolor{mylightblue}\underline{58.40}
      & \cellcolor{mylightblue}\underline{67.51}
      & \cellcolor{mylightblue}66.20
      & \cellcolor{mylightblue}\underline{67.90}
      & \cellcolor{mylightblue}\textbf{54.10} & \;\cellcolor{mylightblue}\underline{71.43} \\
    \midrule
    \multirow{3}{*}{Qwen2.5-VL-72B}
    & Direct & \underline{95.22} & 98.48 & 95.39 & 80.94 & \textbf{77.20}  & 69.16 &65.17 & \underline{96.00} &\underline{99.40} & \;86.33 \\
    & Cascaded & \textbf{95.30}
      & \underline{98.57} 
      & \textbf{96.59} 
      & \underline{82.94} 
      & \underline{75.80}
      & \underline{70.81}
      & \textbf{70.44}
      & \textbf{97.20}
      & \textbf{99.60} & \;\textbf{87.47} \\
    & \cellcolor{mylightblue}CoT &
     \cellcolor{mylightblue}95.07 & \cellcolor{mylightblue}\textbf{98.65} & \cellcolor{mylightblue}\underline{95.65} & \cellcolor{mylightblue}\textbf{83.33} & \cellcolor{mylightblue}\textbf{77.20} & \cellcolor{mylightblue}\textbf{71.57} & \cellcolor{mylightblue}\underline{68.36}&\cellcolor{mylightblue}95.50 &\cellcolor{mylightblue}\underline{99.40} &\;\cellcolor{mylightblue}\underline{87.19} \\
      
    \bottomrule
  \end{tabular}
\end{adjustbox}
  \caption{Results of MLLMs for our three tasks, under the direct and cascaded inferences, as well as the CoT inference (mitigation strategy). Best values per dataset in \textbf{bold}, second-best \underline{underlined}. For all MLLMs, except for Molmo-7B, Cascaded > CoT > Direct, which shows CoT mitigates the skill composition gap, but does not perform as well as cascaded.}
   \label{tab:CoT}
\end{table*}

Although this mitigation strategy performs relatively well, showing improvements over the direct setup in most datasets, it is still far from the performance achieved by the cascaded setup in nearly all cases. Furthermore, the approach is not scalable to arbitrary tasks, as it requires designing prompts manually that explicitly specify which skills the model should apply and in what order. Also, as shown in Appendix~\ref{app:CoT_scale} prompt design itself is a challenging process, with inconsistent effectiveness across different models with distinct prompts. This reinforces the idea that prompt engineering is non-trivial and model-sensitive, highlighting the need to explore alternative strategies for achieving effective skill composition.

Moreover, to determine whether the performance improvements observed in Table~\ref{tab:CoT} are due to more effective skill usage with CoT prompting, we manually analyze the generations of two models on two datasets (more details in Appendix~\ref{app:CoT_output}). The analysis shows that CoT prompting enables MLLMs to compose skills as requested, whereas direct inference shows a different strategy to solve the same tasks. Thus, the performance improvements we observe in Table \ref{tab:CoT} can be attributed to a better skill usage with CoT prompting.

\subsection{Fine-tuning on a cross-modal task} \label{ssec:finetune}

\begin{table*}[htb]
\centering
  \small
  \begin{adjustbox}{max width=\textwidth}
  \begin{tabular}{p{1.5cm} p{1cm}ccccc cc cc }
    \toprule     
      \textbf{Ft data} & \textbf{Setup}& \multicolumn{5}{c}{\textbf{Task 1}} & \multicolumn{2}{c}{\textbf{Task 2}} & \multicolumn{2}{c}{\textbf{Task 3}} \\
      \cmidrule(rl){3-7} 
      \cmidrule(rl){8-9} 
      \cmidrule(rl){10-11}
      &   & \textbf{GSM8K} & \textbf{ARC-E} & \textbf{ARC-C} & \textbf{MMLU} & \textbf{MATH} &\textbf{CV-B} &\textbf{COCO} &  \textbf{Sort}  & \textbf{Sum}\\
    \midrule 
    \multirow{2}{*}{None}
    & Dir. & 63.59 & 69.23 & 61.43 & 46.37 & 36.40&61.55 & 60.00&44.60&11.30  \\
    &  Cas. & 81.05 & 94.74 & 87.88 & 68.21 & 40.40&36.67 & 29.48&57.60&44.20  \\
    
   \midrule \midrule
    GSM8K & Dir. &66.87 \deltapos{3.28}&
    71.04 \deltapos{1.81}&
    63.40 \deltapos{1.97}&
    47.48 \deltapos{1.11}&
    32.80 \deltaneg{3.60}&
    \textbf{64.85} \deltapos{3.30}&
    \textbf{60.26} \deltapos{0.26}&
    46.70 \deltapos{2.00}&
    11.60 \deltapos{0.30} \\ 
    ARC-E/C & Dir. &66.94 \deltapos{3.35}&
    79.88 \deltapos{10.65}&
    65.10 \deltapos{3.67}&
    50.08 \deltapos{3.71}&
    34.80 \deltaneg{1.60}&
    \textbf{64.72} \deltapos{3.17}&
    \textbf{60.90} \deltapos{0.90}&
    45.70 \deltapos{1.00}&
    9.70 \deltaneg{1.60} \\ 
    MATH & Dir. &64.97 \deltapos{1.38} &
    73.48 \deltapos{4.25}&
    60.58 \deltaneg{0.85}&
    46.27 \deltaneg{0.10}&
    38.80 \deltapos{2.40}&
    \textbf{64.21} \deltapos{2.66}&
    \textbf{59.45} \deltaneg{0.55}&
    47.20 \deltapos{2.50}&
    8.90 \deltaneg{2.40}   \\ 
    COCO & Dir. &57.84 \deltaneg{5.75}&
    64.94 \deltaneg{4.29}&
    58.19 \deltaneg{3.24}&
    43.73 \deltaneg{2.64}&
    33.00 \deltaneg{3.40}&
    \textbf{46.44} \deltaneg{15.11}&
    \textbf{69.09} \deltapos{9.09}&
    49.70 \deltapos{5.00}&
    1.30 \deltaneg{10.00} \\ 
    Sort & Dir. &68.23 \deltapos{4.64}&
    69.53 \deltapos{0.30}&
    61.60 \deltapos{0.17}&
    46.47 \deltapos{0.10}&
    38.00 \deltapos{1.60}&
    \textbf{64.34} \deltapos{2.79}&
    \textbf{60.27} \deltapos{0.27}&
    \textbf{63.30} \deltapos{18.60}&
    14.40 \deltapos{3.10} \\ 
    Sum & Dir. &66.18 \deltapos{2.59}&
    71.51 \deltapos{2.28}&
    62.20 \deltapos{0.77}&
    47.12 \deltapos{0.75}&
    34.60 \deltaneg{1.80}&
    \textbf{65.74} \deltapos{4.19}&
    \textbf{60.98} \deltapos{0.98}&
    \textbf{60.60} \deltapos{15.90}&
    \textbf{71.90 }\deltapos{60.60} \\
    \bottomrule
      \end{tabular}
    \end{adjustbox}
    \caption{Results of the Llama3.2 11B model for all tasks under the direct setup when fine-tuning in each dataset. Results of the original model under direct (Dir.) and cascaded (Cas.) inferences are also presented as reference (block tagged as "None"). Numbers in parentheses indicate the difference with respect to the direct baseline. Results that surpass the cascaded baseline are in bold.}
   \label{tab:ft}
\end{table*}

We also explored training strategies for skill composition. More concretely, we have sought to answer the following question: 
(1) how does fine-tuning on a task that requires cross-modal skill composition help reduce the skill composition gap on the same task? and (2) to what extent does this mitigation strategy generalize to different tasks that also require cross-modal skill composition?

To this end we evaluated the full matrix of fine-tuning in each dataset and evaluating on all others. Specifically, we fine-tuned the Llama 3.2 11B model since it achieves good results and shows a significant gap between direct and cascaded inference, while not being the lowest or highest performing model. For fine-tuning, we followed the approach of \citet{chu2025sftmemorizesrlgeneralizes} as a reference and fine-tuned on all datasets except MMLU and CV-Bench since they lack training splits (more details in Appendix~\ref{app:hardware}).%

The goal of our fine-tuning is to reduce or remove the cross-modal skill-composition gap, i.e. we want the model with the visual input (the Direct setup) to perform as well as with the textual input (the Oracle upper bound).  To this end, we rendered each training dataset into images and fine-tuned the MLLM directly on these visual inputs. However, the textual part of the task requires reasoning capabilities, and the selected datasets do not include explicit textual reasoning annotations. To address this limitation, we proceed as follows: (i) we prompt the text-only version of the model with the textual description of the problem, (ii) we store the generated solutions, and (iii) we pair the corresponding image inputs with these generated solutions to construct a new training dataset. 

This approach effectively trains the MLLM to achieve comparable performance when conditioned on visual inputs as when conditioned on textual descriptions, and has the advantage of training on in-distribution model outputs. We chose to fine-tune on the natural text-only model distribution since it preserves the model’s existing reasoning behavior. This methodology allows for studying whether the skill composition gap is reduced. In contrast, rejection-based fine-tuning has the potential to improve the model's reasoning ability, which could result in confounding the analysis.

As shown in Table~\ref{tab:ft}, fine-tuning and evaluating on the same dataset improves direct inference performance. However, this improvement is not sufficient to close the skill composition gap for Task 1, since none of the fine-tuned models surpasses the cascaded inference results of the original model. Analyzing Task 2 results, we observe that most of the fine-tunings outperform the cascaded inference, suggesting that the mitigation strategy pays off. However, it is worth noting that Llama was not originally trained for object detection and counting; therefore, this comparison is not entirely fair (the original model's performance is extremely low). For Task 3, some fine-tunings improve the performance of the model, but only in-distribution training outperforms the cascaded results. 

Our results show that skill-composition-specific fine-tuning has notable generalization capabilities beyond the training dataset. We observe improved performance on other datasets within the same task, indicating a successful transfer of task-specific skill composition. Additionally, performance gains extend to datasets from entirely different tasks, suggesting that fine-tuning enhances the model's broader skill composition capabilities.

Overall, while fine-tuning consistently improves performance across diverse evaluation scenarios, the skill composition gap is only fully closed when fine-tuning and evaluation are performed on the same dataset.

\section{Conclusions}
In this work, we have studied whether MLLMs are able to optimally compose the skills they acquire during pretraining to solve new tasks. To this end, we introduce three multimodal tasks that can be solved by sequentially composing one visual and one textual skill, in a way that would be trivial to find for a human. We compare the model's natural performance with a cascaded approach where we manually force the skill composition, and measure the resulting \textbf{cross-modality skill composition gap}. 

We find that, somewhat strikingly, a significant skill composition gap exists across almost all measured models and tasks, meaning that MLLMs are not able to optimally compose their existing multimodal abilities to solve a task, even in this scenario where the required skill composition is almost trivial. Having shown the existence of this skill composition gap, we explore two possible mitigation strategies: a special CoT that explicitly encourages the required skill composition, and a fine-tuning training strategy that enhances skill-composition, but none of them fully bridge the gap with cascaded inference and present different disadvantages. %

In conclusion, this work prompts further research into why MLLMs have trouble successfully composing skills across modalities even in simple scenarios, as well as into new training approaches for more robust skill composition and generalization.

\section*{Limitations}\label{app:limitations}
Our work currently focuses on the combination of only two skills. However, it would be interesting to incorporate a wider range of skills, both visual and textual, into the tasks. This would be more similar to a real-word scenario in which several skills are required to solve complex problems. Additionally, our work is limited to the evaluation of open-source models. Thus, evaluating commercial models could also be interesting. However, our current evaluation is limited to open-source models because their training data and design choices are reported, making it possible to infer which skills they have likely acquired. This is crucial for accurately assessing skill composition across modalities. While commercial models such as GPT-4~\cite{openai2024gpt4technicalreport}, Gemini~\cite{team2023gemini}, Reka~\cite{team2024reka} or Claude\footnote{\url{https://www.anthropic.com/claude/sonnet}} could offer further insights, interpreting the results meaningfully is challenging due to the limited information available about their training data and learned skills.

\section*{Ethics Statement}
To the best of our knowledge, this paper does not raise any ethical concerns.

\section*{Acknowledgments}
This work has been partially supported by the Basque Government (Research group funding IT1570-22) and the Spanish Ministry of Science, Innovation and Universities (DeepKnowledge project PID2021-127777OBC2, Molvi project PID2024-157855OB-C32, HumanAIze project AIA2025-163322-C61 and  Geo-R2-LLM project PCI2025-163286 by MICIU/AEI/10.13039/501100011033 and co-financed by the European Union). Paula Ontalvilla holds a PhD grant from the Basque Government (PRE\_2024\_1\_0141).

\nocite{*}
\section{Bibliographical References}\label{sec:reference}

\bibliographystyle{lrec2026-natbib}
\bibliography{lrec2026-example}

\bibliographystylelanguageresource{lrec2026-natbib}

\newpage

\appendix
\section{Implementation details}\label{app:implementation}

\subsection{Evaluation}

We evaluated both multiple-choice and open-ended generation datasets using the transformers library of Hugging Face\footnote{\url{https://huggingface.co/}}. For the multiple-choice question datasets (ARC-E, ARC-C, MMLU, and CV-Bench), we explicitly prompted the model to select one of the possible answer choices, as shown in Appendix~\ref{app:prompts}. To extract the model’s selected answer, we applied various regular expressions to identify the predicted letter. Finally, before computing exact-match accuracy, both predicted and ground-truth answers were normalized by converting to lowercase and removing leading or trailing whitespace.

For the open-ended generation datasets, we employed different extraction strategies. In the case of GSM8K, we followed the approach proposed by \cite{eval-harness}, using a combination of filters and regular expressions to extract the final answer. For the remaining datasets (MATH, COCO-Count, Sort, and Sum) we also relied on regular expressions, with dataset-specific filtering functions designed to best extract the model’s generated answer in each case. More details and the specific regex expressions can be found in the project's repository\footnote{\url{https://github.com/paulaonta/skillComposition}}. As with the multiple-choice datasets, exact-match accuracy was computed after normalizing both the predicted and ground-truth answers.

However, for the MATH dataset, we adopted an LLM-based evaluation, in line with a widely-used library\footnote{\url{https://github.com/openai/simple-evals}}. Since both predictions and gold answers often contain \LaTeX{} expressions and certain expressions can be written in different but mathematically equivalent ways (e.g., \lstinline|\frac{2}{3}| and \lstinline|2/3|), string matching is not always a reliable metric. To address this, we used the Reka-Flash~\cite{team2024reka} model for evaluation. This choice was supported by an empirical study in which we measured the agreement between Reka-Flash and human-annotated labels on 200 instances, achieving a 94.5\% match.

\subsection{Hardware information} \label{app:hardware}
Each experiment was not very compute-intensive. All experiments were conducted on an internal cluster equipped with NVIDIA A100-SXM4-80GB GPU. All the experiments were carried out in one A100-SXM4-80GB GPU. However, due to the requirement of a batch size of one, each experiment took between 10 hours and 2 days to complete, depending on the setup. The cascaded setup was the most time-consuming, as it involves a two-step approach.

For the fine-tuning we used 4 NVIDIA A100-SXM4-80GB GPUs with a batch size of 2, gradient accumulation of 16, a learning rate of 1e-6, no LoRA, and a dropout rate of 0.1.

\section{Prompts} \label{app:prompts}
We next described all the prompts used in the experiments. For the design of the prompts we used as a reference different works~\cite{deitke2024molmo, eval-harness, chu2025sftmemorizesrlgeneralizes} and libraries\footnote{\url{https://github.com/openai/simple-evals}}.

\subsection{Prompts for Task 1}
For the evaluation of Task 1, we used different prompts depending on the setup and dataset. Table~\ref{tab:task1_prompts_oracle} presents the prompts used in the oracle setup for each dataset; in this case, the prompt is provided along with the question text. Table~\ref{tab:task1_prompts_direct}, in contrast, shows the prompts used in the direct setup, where both the image and the prompt are given as input. Table~\ref{tab:task1_prompts_cascade} shows the prompts used in the cascaded setup. These prompts correspond to the first step, where the model is asked to extract OCR text from the image. For the second step, we reused the same prompts from the oracle setup (i.e., Table~\ref{tab:task1_prompts_oracle}); in this case, the prompt is provided together with the OCR response extracted in the first step.  Finally, the prompts used in the CoT mitigation strategy are presented in Table~\ref{tab:task1_prompts_COT}.

\subsubsection{Prompts for Task 2}

For the evaluation of Task 2, we used different prompts depending on the setup. In the case of the oracle setup, we only evaluated COCO-Count; the prompt used can be seen in Figure~\ref{fig:task2_prompts_oracle}. Regarding the direct setup, Table~\ref{tab:task2_prompts_direct} shows the prompts for each dataset. However, in the cascaded setup, the prompt was model-dependent because we asked in the first step for the visual skill of object detection, which not all models were trained for. The prompts used in this setup are shown in Table~\ref{tab:task2_prompts_cascade}. For example, since Molmo models have been reported to be trained for pointing tasks, we asked it to point to objects as described in the report~\cite{deitke2024molmo}. For other models, we requested object detection along with their corresponding bounding boxes. For the second step, we reused the same prompts from the oracle setup (i.e., Figure~\ref{fig:task2_prompts_oracle}) provided together with the extracted objects detected in the first step. Finally, the prompts for the CoT mitigation strategy can be seen in Table~\ref{tab:task2_prompts_COT}.

\subsection{Prompts for Task 3}
For the evaluation of Task 3, we followed the same approach as in Task 1. Table~\ref{tab:task3_prompts_oracle} presents the prompts used in the oracle setup for each dataset, while Table~\ref{tab:task3_prompts_direct} and Table~\ref{tab:task3_prompts_cascade} show the prompts used in the direct and cascaded setups, respectively. The prompts for the  cascaded one correspond to the first step, where the model is asked to identify the cards by color, number, and suit. For the second step, we reused the same prompts from the {oracle} setup (i.e., Table~\ref{tab:task3_prompts_oracle}); in this case, the prompt is provided together with the extracted cards in the first step. Finally, the prompts used in the CoT mitigation strategy are presented in Table~\ref{tab:task3_prompts_COT}.

\begin{table*}[htpb]
    \centering
   
    \begin{tabular}{@{}p{1.5cm}p{12cm}@{}}
        \toprule
         \textbf{Dataset} & \textbf{Prompt} \\
        \midrule
         \textbf{GSM8K} & \colorbox{mylightblue}{\texttt{<question>}} Solve the mathematical problem. Provide your reasoning and give a final answer. Conclude your response on a new line with: \texttt{The final answer is [answer]}, where \texttt{[answer]} is the numerical solution to the problem. \\ \midrule
         \textbf{ARC-E}, \textbf{ARC-C}, \textbf{MMLU} & \colorbox{mylightblue}{\texttt{<question>}} Answer the multiple-choice question. The last line of your response should follow this format: \texttt{Answer: \$LETTER} (without quotes), where \texttt{\$LETTER} is one of ABCD. Think step by step before answering. \\ \midrule
        \textbf{MATH} & \colorbox{mylightblue}{\texttt{<question>}} Solve the given math problem step by step. The last line of your response should be of the form Answer: \texttt{\$ANSWER} (without quotes) where \texttt{\$ANSWER} is the answer to the problem. Remember to put your answer on its own line after \texttt{``Answer:''}, and you do not need to use a \texttt{\textbackslash \textbackslash boxed} command.  \\
        \bottomrule
        \end{tabular} 
         \caption{Prompts used in the oracle setup for Task 1.}
    \label{tab:task1_prompts_oracle}
\end{table*}      

\begin{table*}[h]
    \centering
    
    \begin{tabular}{@{}p{1.5cm}p{12cm}@{}}
        \toprule
         \textbf{Dataset} & \textbf{Prompt} \\
        \midrule
        \textbf{GSM8K} & \colorbox{mylightblue}{\texttt{<img>}} Solve the mathematical problem of the image. Provide your reasoning and give a final answer. Conclude your response on a new line with: \texttt{The final answer is [answer]}, where \texttt{[answer]} is the numerical solution to the problem. \\
        \midrule
         \textbf{ARC-E}, \textbf{ARC-C}, \textbf{MMLU} &\colorbox{mylightblue}{\texttt{<img>}} Answer to the multiple choice problem of the image. The last line of your response should follow this format: \texttt{Answer: \$LETTER} (without quotes), where \texttt{\$LETTER} is one of ABCD. Think step by step before answering. \\
        \midrule
         \textbf{MATH} & \colorbox{mylightblue}{\texttt{<img>}} Solve the math problem of the image step by step. The last line of your response should be of the form Answer: \texttt{\$ANSWER} (without quotes) where \texttt{\$ANSWER} is the answer to the problem. Remember to put your answer on its own line after \texttt{``Answer:''}, and you do not need to use a \texttt{\textbackslash \textbackslash boxed} command. \\
        \bottomrule
        \end{tabular} 
        \caption{Prompts used in the direct inference setup for Task 1.}
    \label{tab:task1_prompts_direct}
\end{table*} 

\begin{table*}[h]
    \centering
   
    \begin{tabular}{@{}p{1.5cm}p{12cm}@{}}
        \toprule
         \textbf{Dataset} & \textbf{Prompt} \\
        \midrule         
        \textbf{GSM8K} & \colorbox{mylightblue}{\texttt{<img>}} Given a math problem image: First, extract the text content, and then, solve the problem. Provide your reasoning and give a final answer. Conclude your response on a new line with: \texttt{The final answer is [answer]}, where [answer] is the numerical solution to the problem. \\ 
        \midrule
        \textbf{ARC-E}, \textbf{ARC-C}, \textbf{MMLU} & \colorbox{mylightblue}{\texttt{<img>}} Given a multiple-choice problem image: First, extract the text content, and then, solve the problem. The last line of your response should be of the following format: \texttt{Answer: \$LETTER} (without quotes) where \texttt{\$LETTER} is one of ABCD. Think step by step before answering. \\
        \midrule
        \textbf{MATH} & \colorbox{mylightblue}{\texttt{<img>}} Given a math problem image: First, extract the text content, and then, solve the problem step by step. The last line of your response should be of the form \texttt{Answer: \$ANSWER} (without quotes) where \texttt{\$ANSWER} is the answer to the problem. Remember to put your answer on its own line after \texttt{``Answer:''}, and you do not need to use a \texttt{\textbackslash \textbackslash boxed} command.  \\
        \bottomrule
        \end{tabular} 
         \caption{Prompts used in the CoT mitigation strategy for Task 1.}
    \label{tab:task1_prompts_COT}
\end{table*} 

\onecolumn
\begin{longtable}{@{}p{1.5cm}p{12cm}@{}}
    \toprule
    \textbf{Dataset} & \textbf{Prompt} \\
    \midrule
    \endfirsthead
    
    \multicolumn{2}{c}{\tablename\ \thetable{} -- continued from previous page} \\
    \toprule
    \textbf{Dataset} & \textbf{Prompt} \\
    \midrule
    \endhead
    
    \midrule
    \multicolumn{2}{r}{Continued on next page} \\
    \endfoot

   \endlastfoot
    
    \textbf{GSM8K} & \colorbox{mylightblue}{\texttt{<img>}} Convert the mathematical problem of the image to text. Create a JSONL file with the following structure: include ONLY the question transcription in the `Question' field. For example: \\
    & \begin{tabular}[t]{@{}l@{}}\texttt{\{}\\
    \texttt{~~~"Question": "James writes a 3-page letter to 2 }\\
    \texttt{~~~ different friends twice a week. How many pages does }\\
     \texttt{~~~ he write a year?" }\\
    \texttt{\}} \end{tabular} \\
    & Respond ONLY with the requested JSON structure and nothing else. \\
    \midrule
    \textbf{ARC-E} \quad \textbf{ARC-C}& \colorbox{mylightblue}{\texttt{<img>}} Convert the multiple-choice question from the image to text. Create a JSONL file with the following structure: include ONLY the question transcription in the `Question' field, and the 4 different options in the 'options' field. Write the letter option and the option. For example: \\
    & \begin{tabular}[t]{@{}l@{}}\texttt{\{}\\
    \texttt{~~~"Question": "What do cells break down to produce energy?", }\\
    \texttt{~~~"options": \{}\\
    \texttt{~~~~~~"A": "food", }\\
    \texttt{~~~~~~"B": "water", }\\
    \texttt{~~~~~~"C": "chlorophyll", }\\
    \texttt{~~~~~~"D": "carbon dioxide" }\\
    \texttt{~~~\}}\\
    \texttt{\}} \end{tabular} \\
    \midrule
    \textbf{MMLU} & \colorbox{mylightblue}{\texttt{<img>}} Convert the multiple-choice question from the image to text. Create a JSONL file with the following structure: include ONLY the question and the wording transcription in the `Question' field, and the 4 different options in the `options' field. Write the letter option and the option like this:\\
    & \texttt{\{}\\
    & \texttt{~~~"Question": "Sample question text here?",}\\
    & \texttt{~~~"options": \{}\\
    & \texttt{~~~~~~"A": "First option",}\\
    & \texttt{~~~~~~"B": "Second option",}\\
    & \texttt{~~~~~~"C": "Third option",}\\
    & \texttt{~~~~~~"D": "Fourth option"}\\
    & \texttt{~~~\}}\\
    & \texttt{\}}\\
     &\\
    & For example (DO NOT RETURN THIS EXAMPLE):\\
    & \texttt{\{}\\
    & \texttt{~~~"Question": "Know the self to be sitting in the chariot, the body to be the chariot, the intellect the charioteer, and the mind the reins. He who has understanding, who is mindful and always pure, indeed reaches that place from whence he is not born again, from the Upanishads, India, circa 600s--400s B.C.E. The excerpt above best reflects which of the following religious propositions?",}\\
    &\\
    & \texttt{~~~"Options": \{}\\
    & \texttt{~~~~~~"A": "Hindus should give up their faith in favor of Buddhism.",}\\
    & \texttt{~~~~~~"B": "Individual responsibility, not priestly authority, brings about spiritual evolution.",}\\
    & \texttt{~~~~~~"C": "One lifetime is enough to extinguish one's soul after death.",}\\
    & \texttt{~~~~~~"D": "One's status in the Hindu caste system cannot be altered."}\\
    & \texttt{~~~\}}\\
    & \texttt{\}}\\
     &\\
    & After seeing an image, provide ONLY the JSON object with the actual content from that image. \\
    \midrule
    \textbf{MATH} & \colorbox{mylightblue}{\texttt{<img>}} Convert the entire math problem from the image to text and return it in the following JSONL format.\\
    & Include the full transcription of the question — this means: \\
    &- Preserve all math notation (e.g., LaTeX like \texttt{\$x\^{}2 + 1\$}) \\
    &- Include any diagrams written in Asymptote (between \texttt{[asy]}...\texttt{[/asy]}) exactly as shown \\
    & - Do NOT summarize, shorten, or skip any parts \\
     &\\
    & Output format: \\
    & \texttt{\{}\\
    & \texttt{~~~"Question": "FULL\_QUESTION\_TEXT"}\\
    & \texttt{\}}\\
     &\\
    & For example (DO NOT RETURN THIS EXAMPLE):\\
    & \texttt{\{}\\
    & \texttt{~~~"Question": "In triangle \$ABC\$, \$AX = XY = YB = BC\$ and the measure of angle \$ABC\$ is 120 degrees. What is the number of degrees in the measure of angle \$BAC\$? [asy] pair A,X,Y,B,C; X = A + dir(30); Y = X + dir(0); B = Y + dir(60); C = B + dir(-30); draw(B--Y--X--B--C--A--X); label("\$A\$",A,W); label("\$X\$",X,NW); label("\$Y\$",Y,S); label("\$B\$",B,N); label("\$C\$",C,E); [/asy]"}\\
    & After seeing an image, provide ONLY the JSON object with the actual content from that image. \\  \bottomrule
    &\\
     \caption{Prompts used in the cascaded inference setup for Task 1.}
    \label{tab:task1_prompts_cascade}
\end{longtable}

\newpage

\begin{figure*}[h]
  \centering
  \begin{tcolorbox}[colback=gray!5!white, colframe=gray!75!black, width=0.9\textwidth]
  These are all instances of \colorbox{mylightblue}{\texttt{<obj>}}: \colorbox{mylightblue}{\texttt{<points/bbox>}}. So, \colorbox{mylightblue}{\texttt{<question>}}. Answer to the question. The last line of your response should be of the form \texttt{Answer: \$ANSWER} (without quotes), where \texttt{\$ANSWER} is the numerical answer to the problem.
  \end{tcolorbox}
  \caption{Prompt used in the oracle setup for Task 2}
  \label{fig:task2_prompts_oracle}
\end{figure*}

\begin{table*}[h]
    \centering
    \begin{tabular}{@{}p{2.5cm}p{11cm}@{}}
        \toprule
        \textbf{Dataset} & \textbf{Prompt} \\
        \midrule
        \textbf{CV-Bench} & \colorbox{mylightblue}{\texttt{<img>}} \colorbox{mylightblue}{\texttt{<question>}} The last line of your response should be of the following format: \texttt{Answer: \$LETTER} (without quotes) where \texttt{\$LETTER} is one of ABCDEF. \\
        \midrule
        \textbf{COCO-Count} & \colorbox{mylightblue}{\texttt{<img>}} \colorbox{mylightblue}{\texttt{<question>}}  \\
        \bottomrule
    \end{tabular}
     \caption{Prompts used in the direct inference setup for Task 2.}
    \label{tab:task2_prompts_direct}
\end{table*}

\begin{table*}[htb]
    \centering
   
    \begin{tabular}{@{}p{2.5cm}p{11cm}@{}}
        \toprule
        \textbf{Model} & \textbf{Prompt} \\
        \midrule         
        \textbf{CV-Bench} & \colorbox{mylightblue}{\texttt{<img>}} \colorbox{mylightblue}{\texttt{<question>}} Follow a two-step approach to answer the question. First, detect all \colorbox{mylightblue}{\texttt{<obj>}} in the image as accurately and comprehensively as possible, including overlapping objects. Second, count all detected \colorbox{mylightblue}{\texttt{<obj>}}. Based on the count and the given criteria, determine the correct answer. The last line of your response should be of the following format: \texttt{Answer: \$LETTER} (without quotes) where LETTER is one of ABCDEF. \\
        \midrule
          \textbf{COCO-Count} & \colorbox{mylightblue}{\texttt{<img>}} \colorbox{mylightblue}{\texttt{<question>}} Follow a two-step approach to answer the question. First, detect all \colorbox{mylightblue}{\texttt{<obj>}} in the image as accurately and comprehensively as possible, including overlapping objects. Second, count all detected \colorbox{mylightblue}{\texttt{<obj>}}. Based on the count and the given criteria, determine the correct answer. The last line of your response should be of the form \texttt{Answer: \$ANSWER} (without quotes), where \texttt{\$ANSWER} is the numerical answer to the problem.. \\
        \bottomrule
    \end{tabular} 
     \caption{Prompts used in the CoT mitigation strategy for Task 2.}
    \label{tab:task2_prompts_COT}
\end{table*}

\begin{table*}[htb]
    \centering
    \begin{tabular}{@{}p{2.5cm}p{11cm}@{}}
        \toprule
        \textbf{Model} & \textbf{Prompt} \\
        \midrule
        \small \textbf{LLaVA 1.6-Mistral-7B, LLaVA 1.6-Vicuna-13B, Llama 3.2 11B, Qwen2.5-VL-7B, Qwen2.5-VL-72B } & \normalsize
        \colorbox{mylightblue}{\texttt{<img>}} You are an advanced object detection model. Your task is to analyze an image and detect all instances of \colorbox{mylightblue}{\texttt{<obj>}} within it. For every object detected, provide information in a strict JSON array format with the following structure:

        \begin{tabular}[t]{@{}p{10.5cm}@{}}
        \texttt{[} \\
        \texttt{\ \ \{} \\
        \texttt{\ \ \ \ "label": "ObjectName",} \\
        \texttt{\ \ \ \ "bounding\_box": [x\_min, y\_min, x\_max, y\_max]} \\
        \texttt{\ \ \},} \\
        \texttt{\ \ \ldots} \\
        \texttt{]} \\
        \\
        Where: \\
        - \texttt{"label"} is the name of the detected object \\
        - \texttt{"bounding\_box"} contains the coordinates \texttt{[x\_min, y\_min, x\_max, y\_max]} representing the rectangle enclosing the object \\
        - All values in the bounding box should be float numbers between 0 and 1, representing normalized coordinates \\
        \\
        The response must contain ONLY the JSON array with no additional text before or after. Do not include explanations, introductions, or conclusions.
        \end{tabular} \\
        \midrule
        \small  \textbf{Molmo-7B, Molmo-72B} & \normalsize
        \colorbox{mylightblue}{\texttt{<img>}} You are an advanced object pointing model. Your task is to analyze an image and point out all instances of \colorbox{mylightblue}{\texttt{<obj>}} within it. Your pointing should be as accurate and comprehensive as possible, identifying objects even if they overlap. \\
        \bottomrule
    \end{tabular}
      \caption{Prompts used in the cascaded inference setup for Task 2.}
    \label{tab:task2_prompts_cascade}
\end{table*}

\clearpage

\begin{longtable}{@{}p{1.5cm}p{12cm}@{}}
    \toprule
    \textbf{Dataset} & \textbf{Prompt} \\
    \midrule
    \endfirsthead
    
    \multicolumn{2}{c}{\tablename\ \thetable{} -- continued from previous page} \\
    \toprule
    \textbf{Dataset} & \textbf{Prompt} \\
    \midrule
    \endhead
    
    \midrule
    \multicolumn{2}{r}{Continued on next page} \\
    \endfoot

    \endlastfoot
    \textbf{Sort} & 
    You are given a sequence of 4 playing cards sorted in ascending order by their numeric value. Note that  A counts as `1'. The values are: \\
    &\\
    &\colorbox{mylightblue}{\texttt{<cards>}} \\
    &\\
    &Now, consider a new card:  \colorbox{mylightblue}{\texttt{<new card>}}\\
    &\\
    &Where should this card be inserted in the sequence so that the final list remains in  ascending  order? \\
    &\\
    &Please answer with the index position (0 to 4), where 0 is before the first card, 1 is between the first and second, and so on, up to 4 which means after the last card. \\
    &\\
    &Also, explain your reasoning. \\
    &\\
    &The last line of your response should be of the form \texttt{Answer: \$ANSWER} (without quotes)  where \texttt{\$ANSWER} is the position of the card.\\
    
    \midrule
    \textbf{Sum} & 
    
    \texttt{[}Task Description\texttt{]} \\
   & You are an expert at counting points in card games. You will receive a set of 4 cards. \\
    &Each card must be used exactly once. Your goal is to calculate the total value of the cards according  to the special scoring rules below: \\
    &\\
    &* Number cards (2–10) have a base value equal to their face value \\
    &* Ace = 1 \\
    &* Spades: add 5 bonus points to their base value \\
    &* Clubs with even numbers: subtract 2 from their base value \\
    &* Red cards: their value is twice the base value \\
    &\\
    &\texttt{[}Example calculation\texttt{]}\\
    &For the cards: Card 1: 2, Clubs, Black \\
    &\hspace{5.8em}Card 2: 7, Spades, Black \\
    &\hspace{5.8em}Card 3: 5, Hearts, Red \\
    &\hspace{5.8em}Card 4: Ace, Diamonds, Red \\
    &- 2 of Clubs: Base value 2, even-numbered club (-2), total = 0 \\
   & - 7 of Spades: Base value 7, spade (+5), total = 12 \\
    &- 5 of Hearts: Base value 5, red card (×2), total = 10 \\
    &- Ace of Diamonds: Base value 1, red card (×2), total = 2 \\
    &Sum: 0 + 12 + 10 + 2 = 24 \\
    &\\
    &\texttt{[}Input\texttt{]} \\
    &Cards: \\
    &\colorbox{mylightblue}{\texttt{<cards>}} \\
    &\\
    &Think step by step before answering. \\
   & The last line of your response should be of the form \texttt{Answer: \$ANSWER} (without quotes) where \texttt{\$ANSWER} is the sum of the cards.\\      \bottomrule
   & \\
    \caption{Prompts used in the oracle setup for Task 3.}
    \label{tab:task3_prompts_oracle}\\
\end{longtable}

\begin{longtable}{@{}p{1.5cm}p{12cm}@{}}

    \toprule
    \textbf{Dataset} & \textbf{Prompt} \\
    \midrule
    \endfirsthead
    
    \multicolumn{2}{c}{\tablename\ \thetable{} -- continued from previous page} \\
    \toprule
    \textbf{Dataset} & \textbf{Prompt} \\
    \midrule
    \endhead
    
    \midrule
    \multicolumn{2}{r}{Continued on next page} \\
    \endfoot

    \endlastfoot

    \textbf{Sort} & 
    \colorbox{mylightblue}{\texttt{<img>}} You are observing 4 playing cards sorted in ascending order by their numeric value.  Note that A counts as `1'. \\
    &\\
    &Now, consider a new card:  \colorbox{mylightblue}{\texttt{<new card>}}\\
    &\\
    &Where should this card be inserted in the sequence so that the final list remains in  ascending  order? \\
    &\\
    &Please answer with the index position (0 to 4), where 0 is before the first card, 1 is between the first and second, and so on, up to 4 which means after the last card. \\
    &\\
    &Also, explain your reasoning. \\
    &\\
    &The last line of your response should be of the form \texttt{Answer: \$ANSWER} (without quotes)  where \texttt{\$ANSWER} is the position of the card.\\
    
    \midrule
    \textbf{Sum} & 
    \colorbox{mylightblue}{\texttt{<img>}}\\
    &\texttt{[}Task Description\texttt{]} \\
    & You are an expert at counting points in card games. You are observing four cards in the image.\\
    &\\
    &Each card must be used exactly once. Your goal is to calculate the total value of the cards according  to the special scoring rules below: \\
    &\\
    &* Number cards (2–10) have a base value equal to their face value \\
    &* Ace = 1 \\
    &* Spades: add 5 bonus points to their base value \\
    &* Clubs with even numbers: subtract 2 from their base value \\
    &* Red cards: their value is twice the base value \\
    &\\
    &\texttt{[}Example calculation\texttt{]}\\
    &For the cards: Card 1: 2, Clubs, Black \\
    &\hspace{5.8em}Card 2: 7, Spades, Black \\
    &\hspace{5.8em}Card 3: 5, Hearts, Red \\
    &\hspace{5.8em}Card 4: Ace, Diamonds, Red \\
    &- 2 of Clubs: Base value 2, even-numbered club (-2), total = 0 \\
   & - 7 of Spades: Base value 7, spade (+5), total = 12 \\
    &- 5 of Hearts: Base value 5, red card (×2), total = 10 \\
    &- Ace of Diamonds: Base value 1, red card (×2), total = 2 \\
    &Sum: 0 + 12 + 10 + 2 = 24 \\
    &\\
    &Think step by step before answering. \\
   & The last line of your response should be of the form \texttt{Answer: \$ANSWER} (without quotes) where \texttt{\$ANSWER} is the sum of the cards.\\     \bottomrule
   & \\
       \caption{Prompts used in the direct inference setup for Task 3.}
    \label{tab:task3_prompts_direct}\\
    
\end{longtable}

\begin{longtable}{@{}p{1.5cm}p{12cm}@{}}    
    \toprule
    \textbf{Dataset} & \textbf{Prompt} \\
    \midrule
    \endfirsthead
    
    \multicolumn{2}{c}{\tablename\ \thetable{} -- continued from previous page} \\
    \toprule
    \textbf{Dataset} & \textbf{Prompt} \\
    \midrule
    \endhead
    
    \midrule
    \multicolumn{2}{r}{Continued on next page} \\
    \endfoot

    \endlastfoot

    \textbf{Sort} & 
    \colorbox{mylightblue}{\texttt{<img>}} You are observing 4 playing cards sorted in ascending order by their numeric value.  Note that  A counts as `1'. First, identify the cards and their values. Then, sort them in ascending order. \\
    &\\
    &Now, consider a new card:  \colorbox{mylightblue}{\texttt{<new card>}}\\
    &\\
    &Where should this card be inserted in the sequence so that the final list remains in  ascending  order? \\
    &\\
    &Please answer with the index position (0 to 4), where 0 is before the first card, 1 is between the first and second, and so on, up to 4 which means after the last card. \\
    &\\
    &Also, explain your reasoning. \\
    &\\
    &The last line of your response should be of the form \texttt{Answer: \$ANSWER} (without quotes)  where \texttt{\$ANSWER} is the position of the card.\\
    \\
    \\
    \textbf{Sum} 
    &\colorbox{mylightblue}{\texttt{<img>}}\\
    &\texttt{[}Task Description\texttt{]} \\
    & You are an expert at counting points in card games. You are observing four cards in the image.\\
    &\\
    &Each card must be used exactly once. First identify the cards, their values and their colors. Then, calculate the total value of the cards according to the special scoring rules below:  \\
    &\\
    &* Number cards (2–10) have a base value equal to their face value \\
    &* Ace = 1 \\
    &* Spades: add 5 bonus points to their base value \\
    &* Clubs with even numbers: subtract 2 from their base value \\
    &* Red cards: their value is twice the base value \\
    &\\
    &\texttt{[}Example calculation\texttt{]}\\
    &For the cards: Card 1: 2, Clubs, Black \\
    &\hspace{5.8em}Card 2: 7, Spades, Black \\
    &\hspace{5.8em}Card 3: 5, Hearts, Red \\
    &\hspace{5.8em}Card 4: Ace, Diamonds, Red \\
    &- 2 of Clubs: Base value 2, even-numbered club (-2), total = 0 \\
   & - 7 of Spades: Base value 7, spade (+5), total = 12 \\
    &- 5 of Hearts: Base value 5, red card (×2), total = 10 \\
    &- Ace of Diamonds: Base value 1, red card (×2), total = 2 \\
    &Sum: 0 + 12 + 10 + 2 = 24 \\
    &\\
    &Think step by step before answering. \\
   & The last line of your response should be of the form \texttt{Answer: \$ANSWER} (without quotes) where \texttt{\$ANSWER} is the sum of the cards.\\      \bottomrule
   & \\
    \caption{Prompts used in the CoT mitigation strategy for Task 3.}
    \label{tab:task3_prompts_COT}\\
\end{longtable}

\begin{table}[h]
    \centering
    \begin{tabular}{@{}p{1.5cm}p{12cm}@{}}
        \toprule
        \textbf{Dataset} & \textbf{Prompt} \\
        \midrule
        \textbf{Sort} & 
        \colorbox{mylightblue}{\texttt{<img>}} You are observing four playing cards in the image. Your task is to identify each card and provide a structured response. \\
        &\\
        & For each card, identify: \\
        & 1. The rank: A (Ace), 2-10 \\
        & 2. The suit: Diamonds, Hearts, Spades, Clubs \\
        &\\
        & Return a valid JSON object, listing the cards in the order they appear in the image. Use the exact structure below: \\
        &\\
        & \begin{tabular}[t]{@{}l@{}}\texttt{\{}\\
        \texttt{~~~"suit": ["suit1", "suit2", "suit3", "suit4"],}\\
        \texttt{~~~"number": ["rank1", "rank2", "rank3", "rank4"]}\\
        \texttt{\}} \end{tabular} \\
        
        \midrule
        \textbf{Sum} & 
        \colorbox{mylightblue}{\texttt{<img>}} You are observing four playing cards in the image. Your task is to identify each card and provide a structured response. \\
        &\\
        & For each card, identify: \\
        & 1. The rank: A (Ace), 2-10 \\
        & 2. The suit: Diamonds, Hearts, Spades, Clubs \\
        & 3. The color: Red (Diamonds, Hearts), Black (Spades, Clubs) \\
        &\\
        & Return a valid JSON object, listing the cards in the order they appear in the image. Use the exact structure below: \\
        & \begin{tabular}[t]{@{}l@{}}\texttt{\{}\\
        \texttt{~~~"suit": ["suit1", "suit2", "suit3", "suit4"],}\\
        \texttt{~~~"number": ["rank1", "rank2", "rank3", "rank4"],}\\
        \texttt{~~~"color": ["color1", "color2", "color3", "color4"]}\\
        \texttt{\}} \end{tabular} \\
        \bottomrule
    \end{tabular}
      \caption{Prompts used in the cascaded inference setup for Task 3.}
    \label{tab:task3_prompts_cascade}
\end{table}

\twocolumn

\section{Analysis for Task 1} \label{app:task1}

\begin{table*}[htb]
  \centering
  \begin{adjustbox}{max width=\textwidth}
  \begin{tabular}{llccccc}
    \toprule
    \textbf{Model} &  \textbf{Metric} & \multicolumn{5}{c}{\textbf{Datasets}} \\
     &  &   \textbf{GSM8K} & \textbf{ARC-E} & \textbf{ARC-C} & \textbf{MMLU} & \textbf{MATH}    \\
    \midrule
    \multirow{2}{*}{LLaVA 1.6-Mistral-7B} & WER & 0.0753 & 0.1331 & 0.2214 & 0.7993 & 1.3482 \\
                                       & CER & 0.0647 & 0.1173 & 0.2017 & 0.6893 & 1.0056 \\
    \midrule
    \multirow{2}{*}{LLaVA 1.6-Vicuna-13B} & WER & 0.0331 & 0.1000 & 0.1161 & 0.4257 & 1.1251 \\
                                      & CER & 0.0268 & 0.0915 & 0.1057 & 0.3537 & 0.9140 \\
    \midrule
    \multirow{2}{*}{Llama 3.2 11B} & WER & 0.0610 & 0.0103 & 0.0086 & 0.0741 & 5.7179 \\
                              & CER & 0.0561 & 0.0095 & 0.0076 & 0.0617 & 4.6294 \\
    \midrule
    \multirow{2}{*}{Molmo-7B} & WER & 0.1392 & 0.1912 & 0.2175 & 0.3412 & 0.4598 \\
                              & CER & 0.1079 & 0.1584 & 0.1830 & 0.2915 & 0.4280 \\
    \midrule
    \multirow{2}{*}{Molmo-72B} & WER & 0.0217 & 0.0555 & 0.0629 & 0.2676 & 0.4119 \\
                              & CER & 0.0112 & 0.0380 & 0.0484 & 0.2193 & 0.3178\\
    \midrule
    \multirow{2}{*}{Qwen2.5-VL-7B} & WER & 0.0071 & 0.0112 & 0.0102 & 0.0864 & 0.3701 \\
                                & CER & 0.0055 & 0.0102 & 0.0098 & 0.0758 & 0.2860 \\
    \midrule
    \multirow{2}{*}{Qwen2.5-VL-72B} & WER & 0.0059 & 0.0319 & 0.0401 & 0.2500 & 0.2937 \\
                                & CER & 0.0049 & 0.0322 & 0.0410 & 0.2417 & 0.2434 \\
    \bottomrule
  \end{tabular}
  \end{adjustbox}
  \caption{OCR performance of MLLMs across benchmarks using Word Error Rate (WER) and Character Error Rate (CER). Lower scores indicate better OCR accuracy.}
  \label{tab:ocr_metrics}
\end{table*}
We report in Table~\ref{tab:ocr_metrics} models' OCR accuracy using Word Error Rate (WER) and Character Error Rate (CER) metrics. WER measures the proportion of \textbf{word} level errors (insertions, deletions, or substitutions) relative to the total number of words in the ground truth, while CER performs the same calculation at the \textbf{character} level. 

These metrics provide complementary insights: CER is more sensitive to minor spelling or formatting errors, whereas WER penalizes any deviation that results in an incorrect word. In practice, WER is often higher than CER, since a single character mistake can cause an entire word to be marked as incorrect. 

The lower the value, the better the performance. However, the output is not always a number between 0 and 1. In particular when there are a high number of insertions (for example, when comparing ``hello world'' and ``hello'') the value can be higher than 1.

It can be seen in Table~\ref{tab:ocr_metrics} that the newest models, Qwen2.5-VL-7B, Qwen2.5-VL-72B, Molmo-7B and Molmo-72B achieve the lowest WER and CER scores. Llama 3.2 11B also performs well across most datasets, with the exception of MATH, where its error rates are higher than 1. In contrast, the LLaVA models exhibit the highest WER and CER values, particularly in the MMLU and MATH datasets, indicating weaker visual text recognition in these cases.

\section{Analysis for Task 2} \label{app:task2}
Table~\ref{tab:task2_results_app} presents the exact-match accuracy for models that were not trained specifically for object detection, which is the case of LLaVA 1.6-Vicuna-13B, LLaVA 1.6-Mistral-7B and Llama 3.2 11B, in Task 2.  As anticipated, the proposed skill composition does not work well for this task since the models do not perform correctly in the visual object detection skill as can be shown in Table~\ref{tab:task2_2results_app}.

\begin{table*}[h]
  \centering
  \begin{adjustbox}{height=3cm }
  \begin{tabular}{llccc}
    \toprule    
     &  & \multicolumn{2}{c}{\textbf{Datasets}}  \\
    \textbf{Model} &  \textbf{Setup} & \textbf{CV-Bench} & \textbf{COCO-Count} \\
    \midrule 
    \multirow{3}{*}{LLaVA 1.6-Vicuna-13B} &  Oracle & N/A & 90.89  \\
    \cmidrule(rl){3-4}
    & Direct& 59.77 & 52.97  \\
    &  Cascaded&  64.47 &  63.22 \\
    \midrule
    \multirow{3}{*}{LLaVA 1.6-Mistral-7B} &Oracle & N/A & 92.18 \\
    \cmidrule(rl){3-4}
    & Direct& 60.91 & 54.68   \\
    &  Cascaded& 59.89  & 64.43  \\
    \midrule
    \multirow{3}{*}{Llama 3.2 11B}&  Oracle & N/A & 91.82 \\
    \cmidrule(rl){3-4}
    & Direct& 61.55 & 60.00 \\
    &  Cascaded&  36.67 & 29.48   \\
    \bottomrule
  \end{tabular}
    \end{adjustbox}
      \caption{Exact-match accuracy of MLLMs that were not trained specifically for object detection in Task 2 under three different setups on each benchmark.}
  \label{tab:task2_results_app}
\end{table*}

\begin{table*}[h]
  \centering
  \begin{adjustbox}{max width=\linewidth }
  \begin{tabular}{llc}
    \toprule    
     \textbf{Model} &  \textbf{Skill} & \textbf{COCO-Count}   \\
    \midrule 
    \multirow{3}{*}{LLaVA 1.6-Vicuna-13B}& Obj. detection (visual) & 65.18 \\
    & Str. counting (textual) & 90.89  \\    
     & Cascaded & 63.22 \\
    \midrule
    \multirow{3}{*}{LLaVA 1.6-Mistral-7B} & Obj. detection (visual)& 63.67  \\
    & Str. counting (textual) & 92.18 \\    
    & Cascaded &  64.43  \\
    \midrule
    \multirow{3}{*}{Llama 3.2 11B}& Obj. detection (visual) & 26.74 \\
    & Str. counting (textual) & 91.82  \\    
     & Cascaded & 29.48 \\
    \bottomrule
  \end{tabular}
  \end{adjustbox}
  \caption{Accuracy of the MLLMs that were not trained specifically for object detection for the visual skill of object detection and the textual skill of string counting, which are involved in Task 2 (object counting). We also provide the cascaded result as reference, as the combination of both skills.}
  \label{tab:task2_2results_app}
\end{table*}

\section{Analysis for Task 3} \label{app:task3}

Table~\ref{tab:card_accuracy_sort} presents the image recognition accuracy of the evaluated models in the task of recognizing cards in the Sort dataset. 

The \textbf{Both} column indicates the percentage of cases in which the model correctly predicted both the card number and suit at the correct index in the sorted sequence. The \textbf{Number} and \textbf{Suit} columns report accuracy when only the card number or suit, respectively, matches the ground truth at the correct index. The final column, \textbf{Non-Critical Card Error}, corresponds to cases where the model fails to return the correct cards, but the output does not negatively affect the final result (e.g., if the ground truth is [1, 4, 6, 7] and the new card is 4 of Hearts, but the model returns [1, 3, 5, 8], the insertion position is still correct even though the specific cards are wrong).

We observe that the most recent models, such as Qwen2.5-VL-7B, Qwen2.5-VL-72B, Molmo-7B and Molmo-72B, achieve near-perfect or perfect performance. Llama 3.2 11B also performs very well, but there is a noticeable gap between its accuracy in Both and its performance on cases where only the position is correct but not the specific card details (Non-Critical Card Error). In contrast, with both LLaVA models the gap is especially pronounced. This suggests that these models struggle with fully accurate card recognition at the correct positions.

On the other hand, Table~\ref{tab:card_accuracy_sum} shows the accuracy of the models on the Sum dataset when correctly predicting only the cards' number, suit, color, or all attributes. Since the cards are unordered, the index of the correctly predicted card does not matter. It is important to note that the correct prediction of all attributes is crucial for solving the task correctly, as all attributes are needed to compute the total card values.

We see a similar pattern, i.e., Qwen2.5-VL-7B, Qwen2.5-VL-72B, Molmo-7B, Molmo-72B and Llama 3.2 11B perform well and both LLaVA models show a very poor card recognition performance.

\begin{table*}[h]
\centering

\begin{adjustbox}{max width=\linewidth }
\begin{tabular}{lcccc}
\toprule
\textbf{Model} & \multicolumn{3}{c}{\textbf{Match at correct index}} & \textbf{Non-Critical Card}  \\
& \textbf{Both} & \textbf{Number} & \textbf{Suit} & \textbf{ Error} \\
\midrule
LLaVA 1.6-Mistral-7B & 8.0 & 54.8 & 1.0 & 69.9 \\
LLaVA 1.6-Vicuna-13B & 2.0 & 27.3 & 7.0 & 78.0 \\
Llama 3.2 11B & 82.2 & 87.8 & 89.1 & 93.9 \\
Molmo-7B & 82.4 & 93.1 & 82.4 & 93.1 \\
Molmo-72B & 99.9 & 99.9 & 100.0 & 100.0  \\
Qwen2.5-VL-7B & 100.0 & 100.0 & 100.0 & 100.0 \\
Qwen2.5-VL-72B & 100.0 & 100.0 & 100.0 & 100.0 \\
\bottomrule
\end{tabular}
\end{adjustbox}
\caption{Accuracy of the models in recognizing cards in the Sort dataset.}
\label{tab:card_accuracy_sort}
\end{table*}

\begin{table*}[h]
\centering
\begin{adjustbox}{max width=\linewidth }
\begin{tabular}{lcccc}
\toprule
\textbf{Model} & \multicolumn{4}{c}{\textbf{Match at any index}} \\
& \textbf{All} & \textbf{Number} & \textbf{Suit} & \textbf{Color} \\
\midrule
LLaVA 1.6-Mistral-7B & 0.0 & 43.3 & 8.9 & 39.4 \\
LLaVA 1.6-Vicuna-13B & 1.0 & 17.9 & 9.2 & 39.4 \\
Llama 3.2 11B & 73.2 & 98.3 & 80.7 & 78.2 \\
Molmo-7B & 77.0 & 97.9 & 89.7 & 89.3\\
Molmo-72B & 99.8 & 100.0 &  99.8 & 99.8\\
Qwen2.5-VL-7B & 100.0 & 100.0 & 100.0 & 100.0 \\
Qwen2.5-VL-72B & 99.9 & 100.0 &  99.9 &  99.9 \\
\bottomrule
\end{tabular}
\end{adjustbox}
\caption{Accuracy of the models in recognizing cards in the Sum dataset.}
\label{tab:card_accuracy_sum}
\end{table*}

\clearpage
\clearpage

\section{Details about the datasets} \label{app:dataset}

\subsection{Source datasets}
We obtained all source datasets from the original sources published by the authors. To the best of our knowledge, all the data sources we use are available for non-commercial use.

\begin{itemize}[itemsep=0.1em, topsep=0.2em]
  \item \textbf{GSM8K}~\cite{cobbe2021gsm8k}: We obtained the dataset from the official HuggingFace page\footnote{\url{https://huggingface.co/datasets/openai/gsm8k}} under a MIT License\footnote{\url{https://github.com/openai/grade-school-math?tab=License-1-ov-file}}.
  \item \textbf{ARC Easy and Challenge}~\cite{clark2018think}: We obtained the data from the official HuggingFace page\footnote{\url{https://huggingface.co/datasets/allenai/ai2_arc}} where they specify that it is under a Creative Commons Attribution Share Alike 4.0 license.
  \item \textbf{MMLU}~\cite{hendrycks2020measuring}: We got the data from the official GitHub project\footnote{\url{https://github.com/hendrycks/test?tab=readme-ov-file}} under a MIT License\footnote{\url{https://github.com/hendrycks/test?tab=MIT-1-ov-file}}.
  \item \textbf{MATH}~\cite{hendrycks2024measuring, lightman2023let}: We obtained the dataset from HuggingFace\footnote{\url{https://huggingface.co/datasets/HuggingFaceH4/MATH-500}} under a MIT License\footnote{\url{https://github.com/openai/prm800k/tree/main?tab=MIT-1-ov-file}}.
  \item \textbf{CV-Bench}~\cite{tong2024cambrian}: We obtained the data from the official HuggingFace page\footnote{\url{https://huggingface.co/datasets/nyu-visionx/CV-Bench}} under Apache-2.0 license\footnote{\url{https://github.com/cambrian-mllm/cambrian?tab=Apache-2.0-1-ov-file}}.
\end{itemize}

The rendered textual datasets (GSM8K, ARC-E, ARC-C, MMLU and MATH) can be found in the project's repository.

\subsection{COCO-Count}\label{app:coco}
We used the 2017 validation split of COCO~\cite{lin2014microsoft} from the official project website\footnote{\url{https://cocodataset.org/\#download}} under a Creative Commons Attribution 4.0 License\footnote{\url{https://cocodataset.org/\#termsofuse}} for several reasons: (1) it is an open-source dataset with annotated bounding boxes for each object, and (2) it does not overlap with the other dataset we used, CV-Bench.

Based on this, we created a dataset in which each image is annotated with a single object instance in the format required by each model. The dataset contains a total of 4,952 instances. Since each image may contain multiple objects, we randomly selected only one object per instance to avoid creating an overly large dataset, while still preserving the overall distribution of object counts per image. In other words, we maintained the distribution of the number of objects to be counted, ensuring that models are not always predicting categories with only a single occurrence.

In the case of \textbf{Molmo}, instead of using bounding boxes, the model relies on a custom HTML-like format to point to objects. The \texttt{points} field refers to the center point of each object, which we computed from the bounding box coordinates. For example, for instance ID \texttt{397133}, the object ``people'' is represented as follows. Since there are two \texttt{x} and two \texttt{y} values, it indicates that there are two people in the image:

\begin{lstlisting}[basicstyle=\ttfamily\footnotesize]
"397133": {
    "category": "people",
    "points": "<point x1="443.365"
    y1="208.73000000000002" 
    x2="31.08"  y2="281.195" 
    alt="people">people</point>",
    "count": 2,
    "filename": "000000397133.jpg"
}
\end{lstlisting}

By contrast, for the other models, we used a JSON format consistent with the structure of their outputs. For the same instance, the annotation would look like this, where each object is associated with a separate bounding box:

\begin{lstlisting}[basicstyle=\ttfamily\footnotesize]
"397133": {
    "count": 2,
    "category": "people",
    "points": [
        {
            "label": "people",
            "bounding_box": [
                388.66,
                69.92,
                498.07000000000005,
                347.54
            ]
        },
        {
            "label": "people",
            "bounding_box": [
                0.0,
                262.81,
                62.16,
                299.58
            ]
        }
    ],
    "filename": "000000397133.jpg"
}
\end{lstlisting}

Therefore, in the case of Molmo, the model's counting ability can be evaluated based on its ability to infer the number of distinct coordinate pairs (i.e., different \texttt{x} or \texttt{y} values). For the other models, performance is measured by their ability to correctly detect and count the number of distinct bounding boxes. More details and the dataset can be found in the project's repository.

\subsection{Card game}\label{app:cards}
To create this dataset, we used the work of~\cite{chu2025sftmemorizesrlgeneralizes} as a reference which has a MIT license. We constructed two datasets, both based on poker cards sourced from the aforementioned work. In all cases, we excluded face cards (Queen, Jack, and King) since they can be interpreted in different ways (e.g., all as 10, or as 11, 12, and 13). This exclusion simplifies the task and ensures consistent interpretation.

The datasets are designed to test different reasoning skills: the \textbf{Sort} dataset focuses on ordinal reasoning, while the \textbf{Sum} dataset tests arithmetic reasoning based on visual card attributes. The dataset can be found in the project's repository.

\vspace{0.5em}
\noindent\textbf{Sort Dataset.} We randomly selected four cards and arranged them in ascending order based on their numerical values. While duplicate numbers are allowed, each card (i.e., unique number-suit pair) appears at most once. When two cards share the same number, their relative order is arbitrary, as suit is not considered in the sorting criteria.

\begin{figure*}[h]
    \centering
    \includegraphics[width=0.8\linewidth]{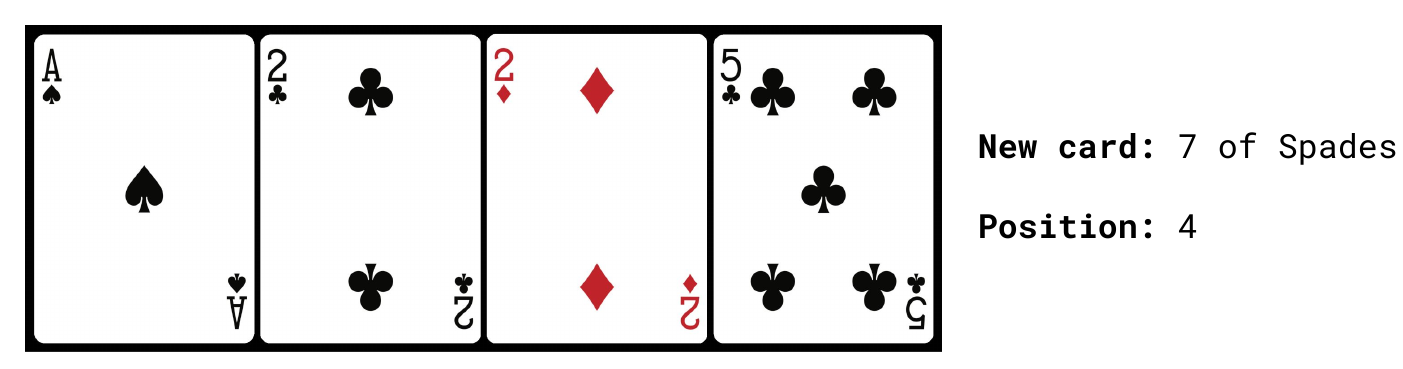}
    \caption{Example from the \textbf{Sort} dataset.}
    \label{fig:sort-example}
\end{figure*}

After forming the initial set of four cards, we randomly selected a fifth card to query. This card was chosen such that it did not appear in the original image and its number did not match any of the cards in the image, to avoid ambiguity. Using this procedure, we created a dataset of 1,000 instances, each consisting of an image showing four sorted cards, a queried card, and the correct position where that card should be inserted. Figure~\ref{fig:sort-example} shows an example.

\vspace{0.5em}
\noindent\textbf{Sum Dataset.} For this dataset, we also randomly selected four cards, but the order is not relevant. As with the Sort dataset, each card is unique in terms of number-suit combination, although duplicate numbers are allowed.

The task is to compute the sum of the four cards using a set of predefined rules, which test the model's ability to recognize different visual features and to and reason about it:

\begin{itemize}
    \item Number cards (2–10) have a base value equal to their face value.
    \item Ace = 1.
    \item Spades: add 5 bonus points to their base value.
    \item Clubs with even numbers: subtract 2 from their base value.
    \item Red cards (Hearts and Diamonds): their value is twice the base value.
\end{itemize}

\begin{figure*}[h]
    \centering
    \includegraphics[width=0.8\linewidth]{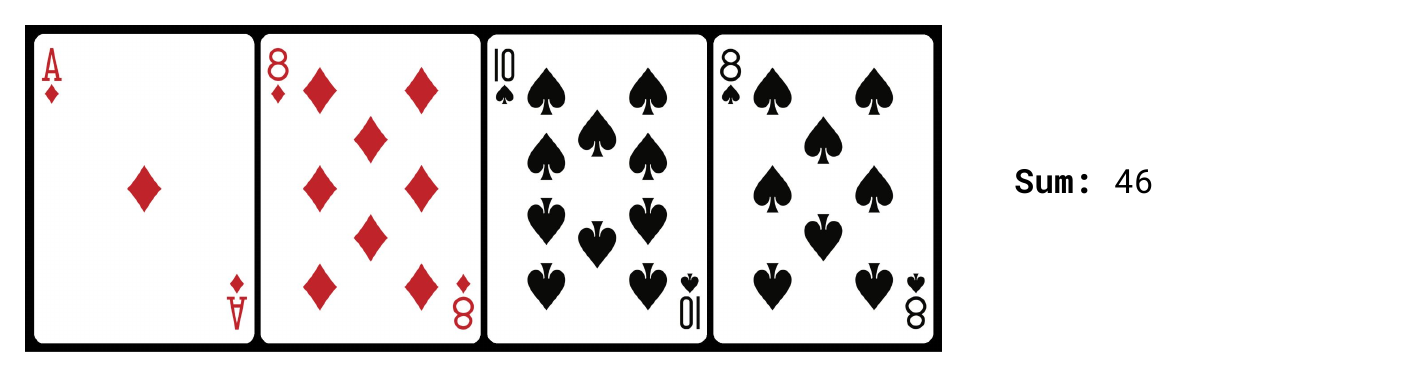}
    \caption{Example from the \textbf{Sum} dataset.}
    \label{fig:sum-example}
\end{figure*}

We generated 1,000 instances, each consisting of an image with four unordered cards and the correct sum based on the rules above. Figure~\ref{fig:sum-example} shows an example.

\section{Composition-specific CoT strategy}\label{app:CoT}

\subsection{Scalability}\label{app:CoT_scale}
To design the CoT mitigation strategy prompts, we employed prompt engineering techniques. While we observed consistent results across all models and datasets, the Sort dataset is an exception.  As shown in Table~\ref{tab:CoT_prompts_results}, we experimented with several CoT prompt  variations (shown in Table~\ref{tab:prompts_try}) for the Sort dataset; however, they did not produce consistent performance gains across all models. 

\subsection{Output analysis}\label{app:CoT_output}
We analyzed some of the model outputs to understand how and why the CoT strategy improves results. This analysis was conducted for the LLaVA 1.6-Mistral-7B model on GSM8K dataset, as it showed one of the largest gains from CoT prompting compared to direct answers. In contrast, we also examined Qwen2.5-VL-7B on CV-Bench dataset, which exhibited the smallest improvement.

In the case of \textbf{LLaVA 1.6-Mistral-7B} we analyzed 50 random samples where the CoT strategy produced correct results, while the direct inference failed. We concluded that CoT prompting does change the behavior of the MLLM, which composes the two skills as instructed. In direct inference, the MLLM uses another strategy, trying to solve the reasoning problem directly over the image. Due to this strategy, we observe that: i) the MLLM commits mistakes when gathering the visual information, and ii) the reasoning step is usually simpler than with the CoT version (see the example~\ref{box:cot_example_llava}). It is worth mentioning that both behaviors are repeated consistently across all the analyzed instances.

For \textbf{Qwen2.5-VL-7B}, we analyzed 66 randomly selected samples where the CoT strategy produced correct results, while the direct inference failed. In this case, we observe again a clear difference between the behavior of the MLLM in direct and CoT setups as it can be seen in \ref{box:cot_example_qwen}. While direct inference tries to analyze the whole image region-by-region, providing short and descriptive captions per image region, CoT prompting makes the model focus only on the target objects and leverages the object detection skill. This again shows that the model is more capable than initially evident.

\begin{table*}[h]
\centering
\begin{tabular}{lcccccccc}\toprule
\textbf{Model} & \textbf{Direct} & \multicolumn{5}{c}{\textbf{CoT}} & \textbf{Cascaded} \\
\cmidrule(lr){3-7}
 &  & \textbf{P1} & \textbf{P2} & \textbf{P3} & \textbf{P4} & \textbf{P5} &  \\
 \midrule
LLaVA 1.6-Mistral-7B & 30.70 & \textbf{38.20} & \textbf{33.40} & \textbf{34.30} & \textbf{38.50} & \textbf{35.30} & 38.40 \\
LLaVA 1.6-Vicuna-13B & 19.70 & 18.70 & \textbf{21.30} & 16.90 & 10.10 & 13.40 & 20.90 \\
Llama 3.2 11B & 44.70 & \textbf{48.30} & \textbf{47.70} & 43.40 & 42.20 & 42.00 & 56.90 \\
Molmo-7B & 42.40 & 26.50 & 31.40 & 39.40 & \textbf{51.30} & \textbf{42.70} & 49.40 \\
Qwen2.5-VL-7B & 58.40 & \textbf{67.50} & 56.20 & 54.60 & 54.00 & 55.90 & 75.50 \\
\bottomrule
\end{tabular}
\caption{Results of the MLLMs with different CoT prompts. In \textbf{bold} appears the results in which there is an improvement respect the direct setup results.}
\label{tab:CoT_prompts_results}
\end{table*}
\onecolumn
\begin{longtable}[h]{@{}lp{13cm}@{}}    
    \toprule
    \textbf{Dataset} & \textbf{Prompt} \\
    \midrule
    \endfirsthead
    
    \multicolumn{2}{c}{\tablename\ \thetable{} -- continued from previous page} \\
    \toprule
    \textbf{Dataset} & \textbf{Prompt} \\
    \midrule
    \endhead
    
    \midrule
    \multicolumn{2}{r}{Continued on next page} \\
    \endfoot

    \endlastfoot

\textbf{P1} & You are observing 4 playing cards sorted in ascending order by their numeric value. Note that A counts as `1'. First, identify the cards and their values. Then, sort them in ascending order.\\
&\\
&Now, consider a new card: \colorbox{mylightblue}{\texttt{<cards>}}\\
&\\
&Where should this card be inserted in the sequence so that the final list remains in ascending order?\\
&\\
&Please answer with the index position (0 to 4), where 0 is before the first card, 1 is between the first and second, and so on, up to 4 which means after the last card.\\
&\\
&The last line of your response should be of the form \texttt{Answer: \$ANSWER} (without quotes) where \texttt{\$ANSWER} is the position of the card. \\
\midrule
\textbf{P2} & You are observing 4 playing cards sorted in ascending order by their numeric value. Note that A counts as `1'. First, identify the cards and their values.\\
&\\
&Now, consider a new card: \colorbox{mylightblue}{\texttt{<cards>}}\\
&\\
&Where should this card be inserted in the sequence so that the final list remains in ascending order?\\
&\\
&Please answer with the index position (0 to 4), where 0 is before the first card, 1 is between the first and second, and so on, up to 4 which means after the last card.\\
&\\
&The last line of your response should be of the form \texttt{Answer: \$ANSWER} (without quotes) where \texttt{\$ANSWER} is the position of the card. \\
\midrule

\textbf{P3} & You are looking at an image showing 4 playing cards sorted in ascending order by their numeric value. First, identify the cards and their values. Note that A (Ace) counts as `1', number cards (2–10) count as their face value.\\
&\\
&Now, consider a new card: \colorbox{mylightblue}{\texttt{<cards>}}\\
&\\
&Where should this card be inserted in the sequence so that the final list remains in ascending order?\\
&\\
&Please answer with the index position (0 to 4), where 0 is before the first card, 1 is between the first and second, and so on, up to 4 which means after the last card.\\
&\\
&The last line of your response should be of the form \texttt{Answer: \$ANSWER} (without quotes) where \texttt{\$ANSWER} is the position of the card. \\
\midrule

\textbf{P4} & You are looking at an image showing 4 playing cards sorted in ascending order by their numeric value. Note that A (Ace) counts as `1'.\\
&\\
&First, identify the cards and their values in the order they appear in the image.\\
&\\
&Now, consider a new card: \colorbox{mylightblue}{\texttt{<cards>}}\\
&\\
&Where should this card be inserted in the sequence so that the final list remains in ascending order?\\
&\\
&Please answer with the index position (0 to 4), where 0 is before the first card, 1 is between the first and second, and so on, up to 4 which means after the last card.\\
&\\
&The last line of your response should be of the form \texttt{Answer: \$ANSWER} (without quotes) where \texttt{\$ANSWER} is the position of the card. \\
\midrule

\textbf{P5} & You are observing 4 playing cards sorted in ascending order by their numeric value. Note that A counts as `1'.\\
&\\
&First, identify the cards and their values in the order they appear in the image. For each card identify: \\ 
&1. The rank: A (Ace), 2–10  \\
&2. The suit: Diamonds, Hearts, Spades, Clubs\\
&\\
&Now, consider a new card: \colorbox{mylightblue}{\texttt{<cards>}}\\
&\\
&Where should this card be inserted in the sequence so that the final list remains in ascending order?\\
&\\
&Please answer with the index position (0 to 4), where 0 is before the first card, 1 is between the first and second, and so on, up to 4 which means after the last card.\\
&\\
&The last line of your response should be of the form \texttt{Answer: \$ANSWER} (without quotes) where \texttt{\$ANSWER} is the position of the card. \\
    \bottomrule
&\\    
    \caption{Prompts used in the CoT mitigation strategy for Task 3.}
    \label{tab:prompts_try}\\
\end{longtable}

\begin{figure}
    \centering
    \begin{tcolorbox}[colback=gray!5, colframe=black!60, boxrule=0.5pt, arc=2mm, left=2mm, right=2mm, title={Example of LLaVA 1.6-Mistral-7B}]

\textbf{Example:} Janet’s ducks lay 16 eggs per day. She eats three for breakfast every morning and bakes muffins for her friends every day with four. She sells the remainder at the farmers' market daily for \$2 per fresh duck egg. How much in dollars does she make every day at the farmers' market?\\

\vspace{0.5em}

\textbf{Direct inference answer:}\\
The image presents a mathematical problem involving a market where a farmer sells eggs and muffins. The farmer has three types of eggs: fresh, medium, and large. The price of fresh eggs is \$2 per dozen, while medium and large eggs are sold at the same price per dozen. The farmer also sells muffins for \$1.50 each.\\

To solve this problem, we need to find the total cost of buying a dozen of each type of egg and one muffin.\\

First, let's calculate the cost of a dozen of each type of egg:\\
- Fresh eggs: \$2 per dozen\\
- Medium and large eggs: \$1.50 per dozen\\

Now, let's calculate the total cost of buying a dozen of each type of egg:\\
- Fresh eggs: \$2\\
- Medium and large eggs: \$3 (two types at \$1.50 each)\\

Add the cost of the muffin:\\
- Total: \$2 + \$3 + \$1.50 = \$6.50\\

The final answer is \colorbox{mylightblue}{\textbf{\$6.50}}.\\

\vspace{0.5em}

\textbf{CoT strategy answer:}\\
The text content of the math problem image is:

"Janet's ducks lay 16 eggs per day. She eats three for breakfast every morning and bakes muffins for her friends every day with four. She sells the remainder at the farmers' market daily for \$2 per fresh duck egg. How much in dollars does she make every day at the farmers' market?"\\

To solve this problem, we need to find out how many eggs Janet has left after eating and baking.\\

Janet eats 3 eggs and uses 4 eggs for baking:\\
3 + 4 = 7 eggs per day\\

Subtract from the total laid:\\
16 - 7 = 9 eggs left to sell\\

Multiply by the price per egg:\\
9 * \$2 = \$18\\

The final answer is \colorbox{mylightblue}{\textbf{\$18}}\\

\vspace{0.5em}

\textbf{True answer:} 18

\end{tcolorbox}
    \caption{Example generation from LLaVA 1.6-Mistral-7B using the direct inference setup and the CoT strategy on the GSM8K dataset.}
    \label{box:cot_example_llava}
\end{figure}

\begin{figure}
    \centering
    \begin{tcolorbox}[colback=gray!5, colframe=black!60, boxrule=0.5pt, arc=2mm, left=2mm, right=2mm, title={Example of Qwen2.5-VL-7B}] 

\textbf{Example:} \\

\includegraphics[width=0.4\textwidth]{img/bottles.pdf} \\[0.25em] 
How many bottles are in the image? Select from the following choices.\\
(A) 0\\
(B) 1\\
(C) 3\\
(D) 2\\

\vspace{0.5em}

\textbf{Direct inference answer:}\\
To determine the number of bottles in the image, let's analyze the objects present:\\

1. There is a table with a decorative item on it that appears to be a decanter or carafe with a stopper, which typically holds liquid.\\
2. There are no other objects that resemble bottles.\\

Based on this analysis, there is only one object that could be considered a bottle.\\

Answer: \colorbox{mylightblue}{\textbf{D}}\\

\vspace{0.5em}

\textbf{CoT strategy answer:}\\
To answer this question, I'll follow the two-step approach:\\

1. Detect all bottles in the image:\\
\hspace{0.5em}- There is a clear glass bottle with a stopper on a small table in front of the chairs.\\

2. Count the detected bottles:\\
\hspace{3em}- Only one bottle is visible in the image.\\

Based on the count, the correct answer is:\\
Answer: \colorbox{mylightblue}{\textbf{B}} \\

\vspace{0.5em}

\textbf{True answer:} B

\end{tcolorbox}
     \caption{Example generation from Qwen2.5-VL-7B using the direct inference setup and the CoT strategy on the CV-Bench dataset.}
    \label{box:cot_example_qwen}
\end{figure}

\end{document}